\documentclass{article}

\usepackage{microtype}
\usepackage{graphicx}
\usepackage{subcaption}
\usepackage{caption}
\usepackage{comment}
\usepackage{booktabs} % for professional tables

\usepackage{hyperref}

\usepackage[accepted]{icml2024}

\usepackage{amsmath}
\usepackage{amsfonts}
\usepackage{amssymb}
\usepackage{mathtools}
\usepackage{amsthm}
\usepackage{comment}

\usepackage[capitalize,noabbrev]{cleveref}

\theoremstyle{plain}
\newtheorem{theorem}{Theorem}[section]
\newtheorem{proposition}[theorem]{Proposition}

\theoremstyle{definition}

\theoremstyle{remark}

\def\e{\mathbf{e}}
\def\x{\mathbf{x}}
\def\y{\mathbf{y}}
\def\u{\boldsymbol{\mu}}
\def\A{\mathbf{A}}
\def\w{\mathbf{W}}
\def\s{\mathbf{s}}

\usepackage[textsize=tiny]{todonotes}

\icmltitlerunning{Predictive Coding beyond Correlations}

\begin{document}

\twocolumn[
\icmltitle{Predictive Coding beyond Correlations}




\begin{icmlauthorlist}
\icmlauthor{Tommaso Salvatori}{verses,wien}
\icmlauthor{Luca Pinchetti}{oxfordcs}
\icmlauthor{Amine M'Charrak}{oxfordcs}
\icmlauthor{Beren Millidge}{oxfordmrc,zyphra}
\icmlauthor{Thomas Lukasiewicz}{wien,oxfordcs}
\end{icmlauthorlist}

\icmlaffiliation{verses}{VERSES Research Lab, Los Angeles, CA 90016, USA}
\icmlaffiliation{wien}{Institute of Logic and Computation, Vienna University of Technology, Austria}
\icmlaffiliation{oxfordcs}{Department of Computer Science, University of Oxford, UK}
\icmlaffiliation{oxfordmrc}{MRC Brain Network Dynamics Unit, University of Oxford, UK}
\icmlaffiliation{zyphra}{Zyphra, Palo Calto, CA, USA}
\icmlcorrespondingauthor{Tommaso Salvatori}{tommaso.salvatori@verses.ai}

\icmlkeywords{Machine Learning, ICML}

\vskip 0.3in
]



\printAffiliationsAndNotice{}  % leave blank if no need to mention equal contribution

\begin{abstract}
Recently, there has been extensive research on the capabilities of biologically plausible algorithms. In this work, we show how one of such algorithms, called predictive coding, is able to perform causal inference tasks. First, we show how a simple change in the inference process of predictive coding enables to compute interventions without the need to mutilate or redefine a causal graph. Then, we explore applications in cases where the graph is unknown, and has to be inferred from observational data. Empirically, we show how such findings can be used to improve the performance of predictive coding in image classification tasks, and conclude that such models are able to perform simple end-to-end causal inference tasks.
\end{abstract}

\section{Introduction}

Predictive coding (PC) is an influential theory of learning and perception in the brain, with roots in Bayesian inference and signal compression \citep{friston2018}. Conventional literature primarily deals with hierarchical models relating top-down predictions to internal states and external stimuli \citep{rao1999predictive}. A recent work went beyond this, and showed how PC can be used to perform inference and learning on structures with any topology, called PC graphs \citep{salvatori2022learning}. Such models are able to perform both discriminative and generative tasks by computing conditional probabilities (i.e., correlations) on Bayesian networks. However, their performance on discriminative tasks are not comparable to the ones obtained by purely hierarchical models. In this work, we first show how PC graphs have the capabilities to go beyond the computation of correlations, and model interventions and counterfactuals \citep{pearl2009causality}. Then, we show how such techniques from causal inference can allow PC graphs reach a performance similar to those of hierarchical models on image classification tasks.

\begin{figure}
  \begin{center}
\includegraphics[width=0.90\linewidth]{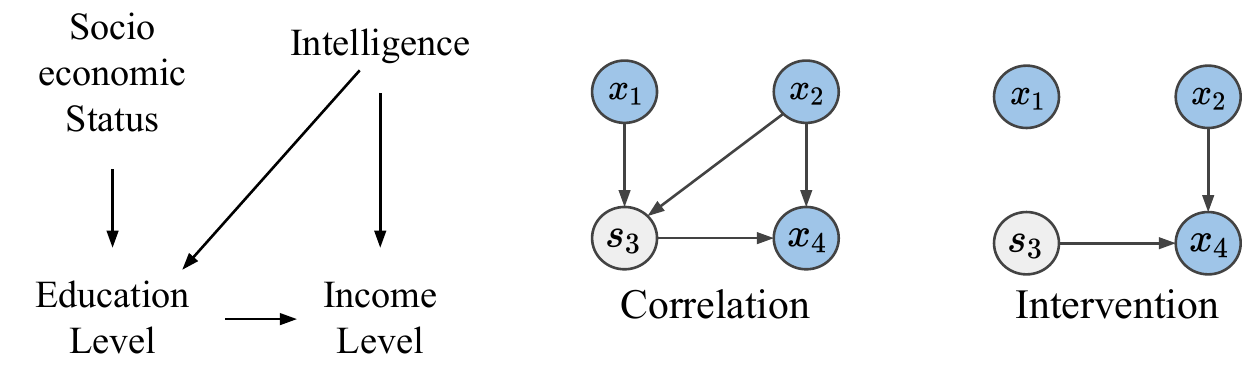}
  \end{center}
  \vspace{-2.5ex}
  \caption{Example socio-economic graph and its structure after conditioning and intervening on \emph{education level}.\vspace{-3ex}}
  \label{fig:causal_graph}
%\vspace{-0.3cm}
\end{figure}

One of the main objects of study in causal inference are \emph{interventions} (or do-operations), usually performed on Bayesian networks by computing a conditional probability on a mutilated graph (see Fig.~\ref{fig:causal_graph}). In Bayesian networks, the process of mutilating the graph can be computationally expensive. A second object of study is structure learning, where the goal is to infer the structure of a graph underlying observational data \citep{zheng2018dags}. The goals of this work are $(i)$ to show both how it is possible to model interventions and perform structure learning in a biologically plausible and efficient fashion, and $(ii)$ to use the developed techniques to improve the performance of PC graphs. Our experiments can then be divided into two categories: the first where we use PC to tackle benchmarks in the causal inference literature, and  the second where we show how interventional queries and structure learning techniques can be used improve the performance of PC graphs in image classification tasks. Our contributions are briefly as follows: \vspace{-2ex}
\begin{itemize}
    \item We introduce \emph{interventional queries}, which  allow to compute interventions on PC graphs efficiently, and without the need of mutilating the graph. Empirically, we test our claims on PC-based structural causal models, and show that the performance is comparable to that obtained by standard baselines. We then show how interventionl queries can be used to improve the test accuracy of  PC graphs on MNIST and FashionMNIST. %\vspace{-1ex}
    \item We show how PC graphs can perform structure learning from observational data. Empirically, we test two techniques that can be used to learn such a structure: using an acyclic prior \citep{zheng2018dags}, or negative examples, the second being a novel contribution. We then show how such techniques further improve the performance of PC graphs on classification tasks.
\end{itemize}\vspace{-2ex}

%We now review the basic concepts of Bayesian networks and PC graphs. Then, we show how the same model can compute interventions by setting the prediction error of a specific node to zero during the inference process. In Section~\ref{sec:sl}, we show how PC graphs can perform structure learning from observational data.
%

\section{Bayesian Networks and Predictive Coding}\label{sec:intro}
Assume a set of $N$ random variables ${\mathbf{X} = \{\x_1,\dots,\x_N\}}$, with $\x_i \in \mathbb R^d$. Relations among variables are represented by a directed graph $G=(V,E)$, also called \emph{Bayesian network} (BN). Every vertex $v_i \,{\in}\, V$ represents  a random variable $\x_i$, and every edge $(i,j)$  a causal relation from $\x_i$ to $\x_j$. Such a BN defines the joint distribution of the system:
\begin{align*}
{  p(\x_1,\dots,\x_N) = \Pi_{i=1}^N \, p(\x_i|par(\x_i)),}
\end{align*}
with $par(\x_i)$ being the parent nodes of $\x_i$. As an example, Fig.~\ref{fig:causal_graph} shows a BN with joint probability
\begin{align*}
{ 
p(\x_1, \x_2, \x_3, \x_4) = p(\x_1)p(\x_2)p(\x_3 \mid \x_1,\x_2)p(\x_4 \mid \x_2,\x_3).}
\end{align*}
Given the graph on the left, conditional probabilities are computed using the full graph structure. Performing an intervention on a specific variable 
($\x_3$ in the figure), on the other hand, requires the computation of a mutilated graph first, obtained by removing all the edges that point to $v_3$. In our example, this reduces to a BN with joint probability 
\begin{align*}
{ 
p(\x_1, \x_2, \x_3, \x_4) = p(\x_1)p(\x_2)p(\x_4 \mid \x_2,\x_3).}
\end{align*}
%
%
%Given the values of specific random variables, we seek to determine the causes and consequences associated with each variable, under the condition of an intervention with all other variables held constant. Here, causes are defined as variables that influence the value of another variable, whereas consequences denote the effects resulting from an intervention on a given variable.
%
More formally, consider the partition $\mathbf{X} = \mathbf{X}_{data} \ \cup \ \mathbf{X}_{unk}$, where the first represents the subset of observed variables $\mathbf{S}_{data}$, while the second represents our random variables that we want to infer. In the example, we would have $\mathbf{X}_{unk} = (\x_1, \x_2, \x_4)$, and $\mathbf{X}_{data} = \x_3$. Our goal is then to infer the posterior distribution 
\begin{align*}
{ 
p(\mathbf{X}_{unk} \ \mid \ \mathbf{X}_{data} =\mathbf{S}_{data}).}
\end{align*}
Such a computation is, however, often intractable. A standard approach to approximate this quantity, is to use variational inference with an approximate posterior $q(\mathbf{X}_{unk})$ restricted to belong to a family of distributions of simpler form than the true posterior.
Then, the goal is now to minimize a KL-divergence between the two distributions. Since the true posterior is not known, one solution is to instead minimize an upper bound on this KL divergence, known as the variational free energy:
\begin{equation}{ 
    F = \mathbb{E}_q[log(q(\mathbf{X}_{unk})) - \log(p(\mathbf{X}_{unk},\mathbf{X}_{data}))].}
\end{equation}
\textbf{Assumptions. \ }We consider each edge $(i,j)$ to be a linear map $\w^{(i,j)}$ composed with a non-linearity $f(\x)$ (e.g., ReLU). This defines how every parent node influences its child nodes. We further set the probability distribution of every node to be a multivariate Gaussian with unitary covariance matrix. In detail, each variable $\x_i$ is sampled from a Gaussian distribution of mean $\mathbf{u}_i$ and variance $1$, where 
\footnote{Every result naturally generalizes to the case of Gaussian distributions with arbitrary covariance matrices $\Sigma$. }
%This, however, will involve the introduction of precisions parameters (inverse covariances). To not make the notation heavier than it already is, we have decided for this easier approach.}
%
\begin{equation}{  
\mathbf{u}_i = \textstyle \sum_{k \in par(i)} \w^{(k,i)}f(\x_k).}
\label{eq:mu}
\end{equation}
The assumption we make on the family of distributions of the approximate posterior are (i) mean field approximation, and (ii) Laplace approximation. The first assumes conditional independence among all variables, while the second assumes they are Gaussians. This is standard in the predictive coding literature, as conditional independence allows every variable to be updated using local information only, while Laplace approximations allow the variational free energy to be a quadratic form \cite{salvatori2023braininspired,friston2003learning, friston2007variational}. From such assumptions, it is possible to derive the following variational free energy:
\begin{equation}{  
F = \textstyle \sum_i \| \x_i - \mathbf{u}_i \|^2 + ln(2\pi).}
\label{eq:energy}
\end{equation}
Such equation is, up to an irrelevant constant, the standard energy function of predictive coding networks, and is what will be used in the remaining of this work to define inference (update of the random variables $\x$) and learning (update of the parameters $\w$). For the exact computations that demonstrate such result, we refer to the original work, and to a recent review \cite{friston2003learning,millidge2020predictive}. When performing causal inference experiments, we will always assume no hidden confounders, and when performing structure learning, we always aim to learn the causal dependencies up to Markov equivalence classes.

\subsection{Predictive Coding Graphs}\label{sec:pcg}

\begin{figure*}[t]
\medskip 
    \centering
    \vspace{-4ex}
	\includegraphics[width=1.0\linewidth]{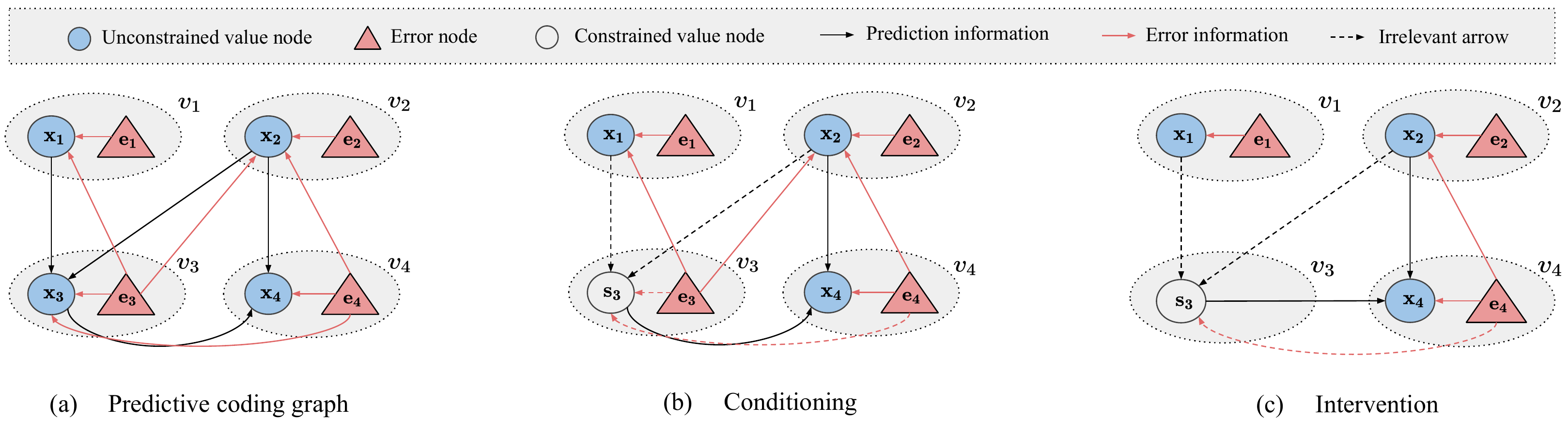}
 \vspace{-3ex}
\caption[short]{$(a)$ PC graph with the same causal structure of that in Fig.~\ref{fig:causal_graph}. Every vertex $v_i$ is associated with a value node $\x_i$, and an error node $\e_i$. The arrows show the influence of every node to the others: the prediction information follows the direction of the arrows of the original graph, while the error information goes backwards. $(b)$ Example of conditioning in PC graphs. We fix the value of $\x_3$, making the effect of all the arrows entering $v_3$ irrelevant, as $\x_3$ is fixed and hence ignores incoming information. This, however, does not apply to error information going \emph{out} from $v_3$, which keeps influencing $\x_1$ and $\x_2$; this is solved in $(c)$ Example of an intervention in PC graphs. According to Pearl's causal theory, the do-operator on a node ($v_3$ in this case) removes the incoming edges, to avoid the newly introduced information to flow backwards and influence the parent nodes. As in PC, the only information flowing opposite to the causal relations is the error information, an intervention can  be performed by removing (or setting to zero) the error node.}
\vspace{-2ex}
\label{fig:scg}
\end{figure*}

As the energy function defined in Eq.~\eqref{eq:energy} is a variational free energy, conditional probabilities can be approximated using PC graphs, a recently introduced flexible, and biologically plausible model \citep{salvatori2022learning}. Each vertex $v_i$ of a PC graph encodes several quantities: the main one is the value of its activity, which changes over time, and represents a random variable. We refer to it as a \emph{value node} $\x_{i,t}$. This is a parameter of the model, which is updated via gradient descent during inference. Additionally, each vertex has a \emph{prediction} $\mathbf{u}_{i,t}$ of its value node, based on input from value nodes of other vertices, as detailed in Eq.~\ref{eq:mu}. The \emph{error} of every vertex at every time step $t$ is then given by the difference between its value node and its prediction, i.e., 
$
\e_{i,t} = \x_{i,t} - \mathbf{u}_{i,t}. 
$
This local definition of error allows PC graphs to learn using only local information. Here, we review the inference phase of PC, which computes correlations among data and results in an approximate Bayesian posterior over all nodes. It is also possible to train these models by updating the parameters $\w$ via stochastic gradient descent over a set of examples. For a detailed description of how \emph{learning} on PC graphs work, see Appendix~\ref{sec:appendix_algo}. 

\textbf{Conditional Query. \ }Assume  a data point $\mathbf{S}_{data} = \{\s_{i_1},$ $\dots,\s_{i_n}\}$. Then, the value nodes $\x_{i_1}, \dots, \x_{i_n}$ of the corresponding vertices are fixed to the entries $\mathbf{S}_{data}$ for every $t$, while the remaining ones are initialized to some random values, and continuously updated until convergence via gradient descent to minimize the energy function, following the rule $ \Delta{\x}_{i,t} \propto {\partial F_t}/{\partial \x_{i,t}}.$
The unconstrained sensory vertices will converge to a minimum of the energy given the fixed vertices, thus computing the conditional expectation of the latent vertices given the observed stimulus. Formally, the inference step estimates the conditional expectation
\begin{equation}{ 
E(\mathbf{X}_T \mid \forall t\colon (\x_{i_1,t},\dots,\x_{i_n,t}) = (\s_{i_1}, \dots, \s_{i_n}) ),\label{eq:cond_query}}
\end{equation}
where $\mathbf{X}_T$ is the matrix of all value nodes at convergence. This computes the \emph{correlation} among different parameters of the graph. In the next section, we show how to model \emph{interventions} in PC graphs. For a neural implementation of a PC graph, see  Fig.~\ref{fig:scg}a, and for the neural implementation of a conditional query, where the value of a specific node is fixed to a data point, see~Fig.~\ref{fig:scg}b.

\vspace{-1ex}
\section{Causal Inference via Predictive Coding}\label{sec:ci}
\vspace{-1ex}

\begin{figure*}[t]
\medskip 
    \centering
	\includegraphics[width=0.9\linewidth]{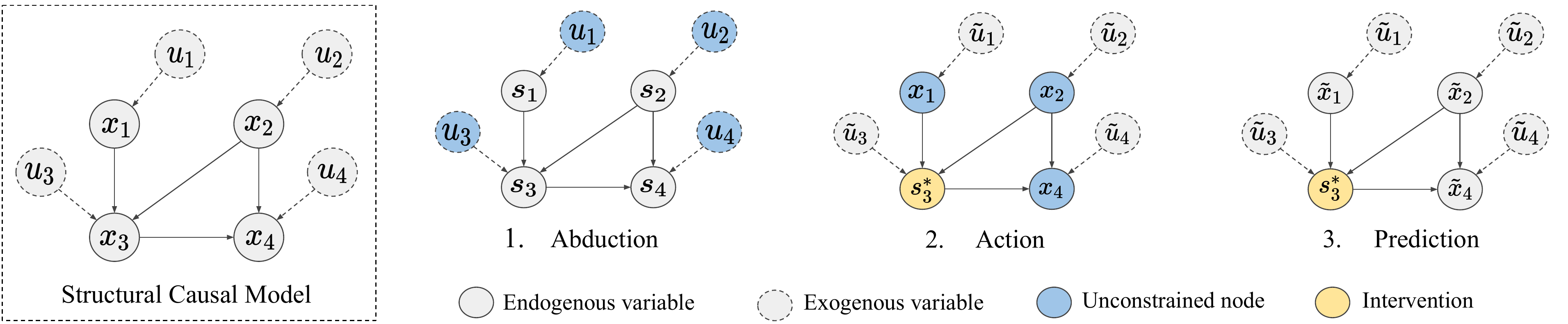}
\caption[short]{\emph{What would $\x_4$ be, had $\x_3$ been equal to $\s^*_3$ in situation $U = \mathbf{u}$?} This figure provides an example of the three-step process to perform counterfactuals, using a structural causal model with four exogenous and four endogenous variables. We are given two kinds of data: the original values of $\x_1,\dots, \x_4$, which correspond to past information, here denoted by $\s_1,\dots, \s_4$, and the intervention information $\s_3^*$, needed to understand the \emph{what would have happened to $\x_4$ if we had changed $\s_3$ to $\s_3^*$?}. The final answer corresponds to the node $\tilde \x_4$ obtained in the prediction step.\vspace{-4ex}}
\label{fig:SCM}
\end{figure*}

The main goal of causal inference is to be able to simulate interventions in a process, and study the counterfactual effects of such interventions. In statistics, \emph{interventions} are denoted by the $do(-)$ operator \citep{pearl1995causal,pearl2009causality}. The value of a random variable $\x_i$ when performing an intervention on a different variable $\x_j$ is denoted by $p(\x_i \mid do(\x_j = \s))$. This is equivalent to the question \emph{What would $\x_i$ be in this environment if we set $\x_j = \s$?} In the case of the example in Fig.~\ref{fig:causal_graph}, the question could be \emph{What would the expected income level be, if we change the education level of this person?} In fact, while `education' and `income level' may be correlated by a hidden confounder (intelligence, in this case), an intervention removes this correlation by changing the education level of a randomly selected individual, regardless of level of intelligence.
%\footnotetext{This is different from correlation: Let us assume that education itself yields no direct impact on income. Nevertheless, we might still discern a correlation between education and income due to a confounding variable influencing both parameters, e.g., intelligence. Here, a robust correlation manifests, but if we were to implement an intervention—--say, increasing the educational attainment of a randomly selected individual—--it might not substantially affect their income.}
To perform an intervention on a Bayesian network, we first have to act on the structure of the graph, and then query the model by conditioning on the new graph, as shown in Fig.~\ref{fig:causal_graph}. Assume that we have a PC graph $G$, and we want to know the value of $\x_i$ after performing an intervention on  $\x_j$ by fixing it to some value~$s$. This can be done according to the two following steps:\vspace*{-1.5ex}
\begin{enumerate}
    \item Generate a new graph $G'$ by removing all the in-coming edges of $v_i$ from $G$.\vspace*{-1ex}    
    \item Perform the conditional query $E(\mathbf{X} \mid  \x_{j} = \s)$ on $G'$.\vspace*{-1.5ex}
\end{enumerate}
We now show that, when dealing with predictive coding graphs, it is possible to perform an operation that is equivalent to an intervention, without the need of mutilating the graph, that is, without performing step $1.$

\textbf{Interventional Query. \ }In a PC graph, the only information that flows in the opposite direction of an arrow is the prediction error. In fact, if we have $v_1 \rightarrow v_2$, the update of the value node $\x_1$ is affected by the error $\e_2$. To avoid this and perform an intervention $p(\x_1 \mid do(\x_2 = \s_2))$, we first set (and fix) the value of $\e_2$ to zero, and then perform a conditional query by setting $\x_2 = \s_2$. This is convenient, since it allows us to not directly act on the structure of the graph to perform interventions but rather perform them dynamically `at runtime', which results in increased efficiency in the case of nodes with numerous incoming edges. We assume hard interventions over soft ones \citep{correa2020calculus}, thus eliminating all parent variable effects. Hence, the following result holds, proven in Appendix~\ref{sec:appendix_proof}:

\begin{proposition}
Let $\mathcal G$ be a PC graph, with structure given by a directed acyclic graph $G$ with variables $\{\x_1, \dots, \x_N\}$, as defined in Sec. 2.1. Then, the distribution of the variables obtained after the following two operations are equivalent:\vspace*{-1ex}
\begin{itemize}
\item A \emph{conditional query} performed by setting $\x_j = \s_j$ of the modified PC graph, $G'$, where $G'$ is derived from $G$ by the excision of all incoming edges to nodes $x_j$. The precise formulation of a conditional query is as articulated in Eq.~\ref{eq:cond_query}.\vspace*{-0.5ex}%
\item An \emph{interventional query} performed by setting $\x_j = \s_j$ on the original PC graph with structure $G$, which involves executing a \textit{conditional query} on the node $\x_j$, while also imposing the condition $\e_{j,t} = 0$ for every time step $t > 0$. That is,
\vspace*{-1ex}%
\end{itemize}
\begin{equation*}{ 
    E(\mathbf{X_T} \mid do(\x_j = \s)) = E(\mathbf{X_T} \mid \forall t: \x_{j,t} = \s, \e_{j,t} = 0 \ \forall t).}
\end{equation*}
\end{proposition}

\subsection{Structural Causal Models via PC Graphs}

While interventions serve to answer questions about the consequences of actions performed in the present, counterfactuals are used to study interventions in the past, e.g., \emph{What would the value of $\x_i$ have been if $\x_j$ had been set to $\s^*_j$, given a particular contex?} This is modeled using structural causal models (SCMs), where the \emph{context} is defined.
An SCM is a triple $(U, V, F)$, where $V$ is the set of endogenous (observable) variables corresponding to the internal vertices of the graph, $U$ is the set of exogenous (unobservable) variables, denoted $\mathbf{\mu}_i$, that serve as root nodes in the graph, and $F$ is the set of functions that determine the values of endogenous variables according to the structure of $G$. An example of an SCM is represented in Fig.~\ref{fig:SCM}. Then, counterfactual inference with an SCM involves three steps:\vspace*{-1ex}%
\begin{enumerate}%[align=left,leftmargin=*]
    \item \textbf{Abduction}: Here, we are provided with the values $(\s_1, \dots, \s_N)$ of the endogenous nodes in $V$. We use them to compute the values of the exogenous variables, which we denote by $\tilde u_1, \dots, \tilde u_N$. Hence, according to 
    \begin{equation*}{ 
    E(\mathbf{\mu}_1, \dots, \mathbf{\mu}_N \mid \forall t: (\x_1, \dots, \x_N) = (\s_1, \dots, \s_N)).}
    \end{equation*}
    \item \textbf{Action}: Now that we computed the values of the exogenous variables, we fix them and perform an intervention on $\x_j$. Particularly, we set $\x_j = \s^*_j$, and we set $\e_j = 0$, which has the effect of removing any influence of $x_j$ on its parent nodes.
    \item \textbf{Prediction}: We now have all the elements to compute the counterfactual on $\x_i$, which is:\vspace*{-1ex}   
    \end{enumerate}
\begin{equation*}{ 
    E(\x_i \mid \forall t: (\mathbf{\mu}_1, \dots, \mathbf{\mu}_M) = (\tilde{\mathbf{\mu}}_1, \dots, \tilde{\mathbf{\mu}}_M), \ \x_j = \s^*_j, \ \e_j = 0).}
\end{equation*}

\vspace{-1ex}
\subsection{Experiments}
\vspace{-1ex}

We now perform experiments that confirm the technical discussion and claims made in the previous section. To this end, we test PC graphs in their ability to compute causal queries on three levels of Pearl's ladder of causation (\citeyear{pearl2009causality}), namely, association (level 1), intervention (level 2), and counterfactual (at level 3) for both linear and non-linear data. Then, we show how interventions can be used to improve the performance of classification tasks on fully connected models. We conclude with an experiment showcasing the robustness of PC graphs and their learned representations of complex data, specifically, with respect to counterfactual interventions on distinct attributes of images. We assume causal sufficiency (no unobserved confounders) and compare against works that make this assumption. The appendix contains extensive results and details to reproduce our approach.
%
%\vspace{-1ex}

\textbf{Causal Inference. }We evaluate associational, interventional, and counterfactual queries on various directed acyclic graphs (Fig.~\ref{fig:all_DAGs}) using linear and non-linear data (Table \ref{tab:nonlinear_data_equations} Appendix~\ref{sec:app_cg}). We assume the graphs are known and the SCMs have additive Gaussian noise. We generate synthetic observational, interventional, and counterfactual data for testing. 
The observational data are generated by randomly sampling values for the exogenous variables, $\mathbf{u}$. We then use the exogenous values to compute the values of the endogenous variables $\x$.
%Interventional data is generated following a similar procedure, with the difference that they use a SCM whose structural equations are altered by interventions.
%A similar procedure is used to generate interventional and counterfactual datasets, with the difference that they use a SCM whose structural equations are altered by interventions.  
Interventional data are similarly generated, but the structural equations are altered by an intervention. 
Finally, the counterfactual data consist of pairs, ($\x, \x'$)
%, for a unique setup of exogenous node values, $\u$,
with $\x$ being observational and $\x'$ interventional data, both sharing the same $\mathbf{u}$.
To perform causal inference, we fit the PC graph to the observed data to learn the parameters of the structural equations, including proxy noise distribution parameters. We evaluate the learned SCM by comparing various difference metrics between true and inferred counterfactual values. Details on all metrics are in Appendix~\ref{sec:appendix_CI}.
%We evaluate the performance of our model in answering queries of all causal hierarchies (association, intervention, counterfactual) for all test sets, and study various error metrics of the estimated distributions between the true query values the inferred ones, such as the Maximum Mean Discrepancy (MMD), Mean Absolute Error (MAE), and Mean Squared Error (MSE).

%Our proposed method does not require more parameters than the number of parameters of the true structural equations. As such, it is lightweight, simple to train, and not prone to overfitting. 

\begin{figure*}[t]
\medskip 
    \centering
	\includegraphics[width=0.95\linewidth]{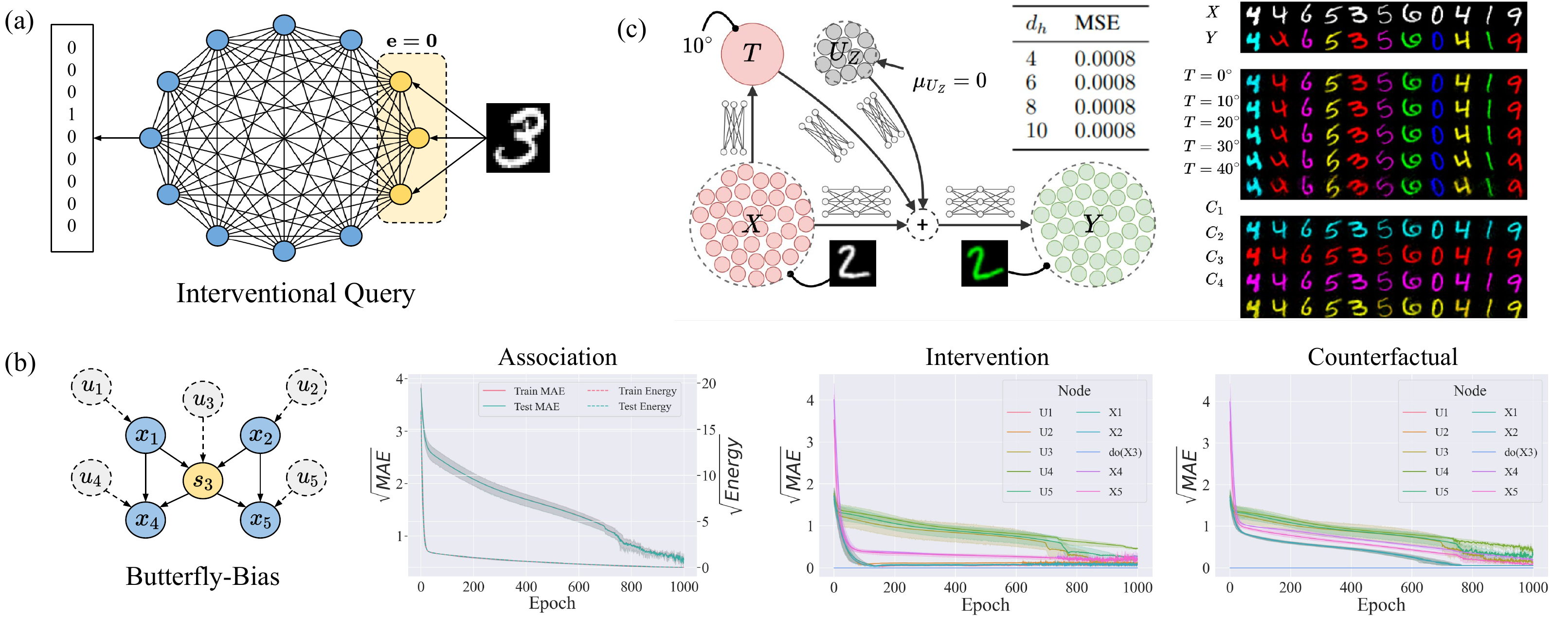}
 \vspace{-2ex}
\caption[short]{(a) How to compute a prediction given a data point on a fully connected PC graph, using interventional queries. 
(b) Left to right: causal structure of the SCM. 
Convergence behavior of PC energy vs. error metric (MAE), during SCM learning for butterfly graph.
Error (by node) of interventional query estimates on $\x_3$ (yellow node). 
Error (by node) of counterfactual query estimates with intervention on $\x_3$ given factual data (blue nodes). (c) Architecture used to reconstruct  counterfactual images. $U_Z$ corresponds to the color of the digit, $T$ to the rotation angle, $X$ to the input, $Y$ to the colored and rotated image.  The architecture represented is a predictive coding network that resembles the architecture described in the main paper, where the clusters of neurons represent nodes. Arrows represent transformations achieved via MLPs. The reconstructions on the right demonstrate the robustness of our model with respect to intervention to (up) rotation angle and (down) color by transforming each input digit to the desired target.The colored digits show that our method is robust when performing interventions on the rotation angle. The table shows that the model performance does not  depend on the choice of the number of neurons $d_h$ for the node $\mathbf{u}_z$. The work proposing the experiment \citep{de2022deep} reports an MSE of $0.001$.}
\vspace{-2.5ex}
\label{fig:caus_exp}
\end{figure*}

Here, we only provide results for the most interesting and complex graph among the proposed ones, namely, the butterfly, represented in Fig.~\ref{fig:caus_exp}(c). We refer the reader to Appendix~\ref{sec:appendix_CI} for a detailed study of all the aforementioned graphs on a large number of metrics. The experimental results show that PC graphs accurately estimate causal queries for linear and more complex non-linear data. The plots in Fig.~\ref{fig:caus_exp}(c) show that the model is able to correctly infer interventional and counterfactual queries, as shown by the converging MAE of non-intervened nodes. Finally, unlike \citep{CAREFL}, we do not reduce the graph to its causal ordering and the performance of PC graphs remains stable as causal paths get longer, an issue seen in \citep{VACA}. More detailed results and a discussion on the common graph causal query experiments are in Appendix \ref{sec:app_cg}.%\vspace{-2ex}
%Furthermore, this method only requires parameters than the number of parameters of the true structural equations. As such, it is lightweight, simple to train, and not prone to overfitting. 

%\smallskip 
\textbf{Classification. \ }In the original work on PC graphs \citep{salvatori2022learning}, the authors have trained a fully connected model to perform classification tasks using conditional queries. The performances are poor compared to those of hierarchical models for two reasons: first, conditional queries do not impose any direction to the information flow, making the graph learn $p(\bf{y}|\bf{x})$ as well as $p(\bf{x}|\bf{y})$, even though we only need the first term. Similarly, the model also learns $P(\x)$ and $P(\bf{y})$, which is, how the prior depends on itself. Second, the model complexity is too high, not allowing the model to use any kind of background knowledge, such as hierarchical/structural information/prior, usually present in sparser models. Here, we address the first limitation by performing an \emph{interventional query} on the input, which is fixing the input nodes to the input values, and zeroing out their errors. This prevents the error of the inputs to spread in the network, and enforces a specific direction of the information flow, which goes from cause (the image) to effect (the label). To assess the impact of such an intervention on the test accuracy, we train a fully connected model with $2000$ neurons on the MNIST and FashionMNIST datasets, and compute the test accuracy for conditional and interventional queries. We perform a large hyperparameter search on learning rates, activation functions, and weight decay values. In all cases, performing an intervention improves the results. In Fig.~\ref{fig:sl_exp}(d), we present an example showcasing the best results obtained after hyperparameter search. The interventional query led to improvements in the final test accuracy of almost $2\%$ for both datasets. The experiment details can be found in Appendix~\ref{sec:appendix_classification}.

%\vspace{-1ex}
%\smallskip 
\textbf{Counterfactual Robustness. \ }A recent work demonstrates that existing deep-learning methods fail to obtain sufficient robustness or performance on counterfactual queries in certain scenarios \citep{de2022deep}. We show that PC surpasses current state-of-the-art results for counterfactual queries while requiring a simpler architecture, and without relying on ad hoc training techniques. We evaluate the robustness of our model on counterfactual inference tasks of higher dimensions, thereby examining the feasibility of our method to perform causal inference on more complex data. The dataset we consider consists of tuples $(\x, \mathbf{u}_z, \mathcal{T}, \y, \mathcal{T}', \y')$, where $\x$ is an image from the MNIST dataset, $\mathcal{T}$ is the assigned treatment, which is a rotation angle (confounded by $\x$). Furthermore, $\mathbf{u}_z$ is a hidden exogenous random variable that determines the color of the observed outcome image $\y$, which is added to the fourth variable $\y'$, a colored and rotated MNIST image representing the counterfactual response obtained when applying the alternative treatment $\mathcal{T}'$. We consider SCMs with $4$ nodes that encode the four variables, as sketched in Fig.~\ref{fig:caus_exp}(c). Here, every edge represents a feed-forward network of different depth and hidden dimension $1024$. A detailed explanation of how to reproduce the results is given in Appendix~\ref{sec:appendix_robustness}.

\vspace{-1ex}
The results show that PC graphs improve on state-of-the-art methods, despite the fact that we do not use convolutional layers like in the original work. First, the generated images have an MSE on the test set ($0.0008 \pm 0.0002$) that is lower than that reported in the original work ($0.001 \pm 0.001$). The high quality of reconstruction is also visible in the generated images  in Fig.~\ref{fig:caus_exp}(c). 
Compared to \citep{de2022deep}, we are able to generalize to rotations of $40^\circ$ (absent in the training data), even if this introduces some noise in the generated output. Furthermore, contrary to the original model, our architecture is robust relative to the choice of the hyperparameter linked to $u_z$ and does not necessitate to perform a hyperparameter sweep to find the right value. So, we conclude that PC graphs are able to correctly model the treatment rotation in the counterfactual outcome, while keeping the color, which is independent of rotation, unchanged.

\vspace{-1ex}
\section{Structure Learning}\label{sec:sl}
\vspace{-1ex}

Learning the causal structure from observational data is a useful process for explainability and modeling interventions. Traditional approaches, which use combinatorial search algorithms, tend to become computationally expensive and slow as the complexity (e.g., the number of nodes)  of the graph increases \citep{chickering1996learning,chickering2002optimal}. Therefore, modern approaches focus on gradient-based learning methods instead \citep{zheng2018dags}, as they allow us to handle larger graph structures in a computationally efficient manner. Let us consider $\A$ to be the adjacency matrix of a graph. Ideally, this matrix should be a binary matrix with the property that $a_{i,j} = 1$, if there exists an edge from $v_i$ to $v_j$, and $0$, otherwise. From a Bayesian perspective, our method learns the marginal of the graph edges, where $\A$ is a matrix composed of continuous, learnable parameters, which assign weights to signify importance of specific connections. To this end, we can consider every PC graph to be fully connected, where the prediction of every node $\x_i$ now depends on the entries of the adjacency matrix:
\begin{align*}
& \mathbf{u}_i = \textstyle \sum_{k=0}^N a_{k,i}f_{k,i} (\x_k),
\\ 
& \mathrm{update \ rule:} \ \ \   \Delta a_{i,j} \propto - {\partial F} /  {\partial a_{i,j}} = \beta\cdot \e_{i,T} \w^{\top} \!\! f ( \x_{j,T}),
%\label{eq:mu_causal}
\end{align*}
where $\beta$ is the learning rate of the parameters  $a_{i,j}$. Then, the entries of $\A$ are updated via gradient descent to minimize the variational free energy of Eq.~\ref{eq:energy}. Our goal is to learn an acyclic, sparsely connected graph, which requires a prior distribution that enforces these two constraints. We consider three possible priors: a Gaussian prior, a Laplace prior, and the acyclicity prior. The latter  is equal to zero iff the corresponding graph is acyclic \citep{zheng2018dags}:
\begin{align*}
& l(\A) = exp( - \textstyle \sum_{i,j} | a_{i,j}|), \\ 
& g(\A) = \mathcal N(0,1), \ \
\\ 
& h(\A) = tr(exp(\A \times \A)) - d.  
\label{eq:priors}
\end{align*}
The energy function to minimize via gradient descent is  the sum of the total energy, as defined in Eq.~\ref{eq:energy}, and the three aforementioned prior distributions, each weighted by a scaling coefficient. The first two priors effectively apply the $L1$ and $L2$ norms to the parameters of the adjacency matrix, and they form the elastic norm when used in conjunction.

\textbf{Negative Examples. \ }The regulariser $h(A)$ introduces an inductive bias that may be undesirable, as we know that cyclic structures may be beneficial in several tasks \citep{salvatori2022learning}. Without $h(A)$, however, training converges towards a degenerate structure, as shown on the top right of Fig.~\ref{fig:sl_exp}(c), where each output node predicts itself, ignoring any contribution of the input nodes. We solve this degeneration behavior by introducing negative examples, which are data points with a wrong label, into the training set. The model is then trained in a contrastive way \citep{simclr}, i.e., by \emph{increasing} the prediction error of every node for negative examples $k$, and decreasing it otherwise, when the label is correct; see Fig.~\ref{fig:sl_exp}(c). A detailed explanation of how training with negative samples works is given in Appendix~\ref{sec:appendix_structure_learning}. We show that negative examples address the convergence issue towards a degenerate graph by rendering the label nodes contingent on the inputs, thus steering the model towards adopting a hierarchical structure instead.

%\paragraph{End-to-end causality engine}
%Models that perform end-to-end causal inference are an active topic of research, due to their ability of both learning the causal structure from observational data, and modeling associational, interventional, and counterfactual distributions \citep{geffner2022deep, sharma2020dowhy}. As PC also allows to perform both tasks sequentially, using the same framework, it is an end-to-end causality engine, that can answer causal queries without knowing the child-parent relationships in the structural equations of the SCM. In the supplementary material, we provide experimental evidence that this is indeed the case by performing end-to-end causal inference tasks on different kinds of graphs. 

%
%\begin{equation}
%F(x,\w,A) = \sum_i \| \e_i \|^2 + \lambda_l l(A) + \lambda_g g(A) + \lambda_h h(A).
%\label{eq:gen_energy}
%\end{equation}

\vspace{-1ex}
\subsection{Experiments}
\vspace{-1ex}

\begin{figure*}[t]
\medskip 
    \centering
	\includegraphics[width=\linewidth]{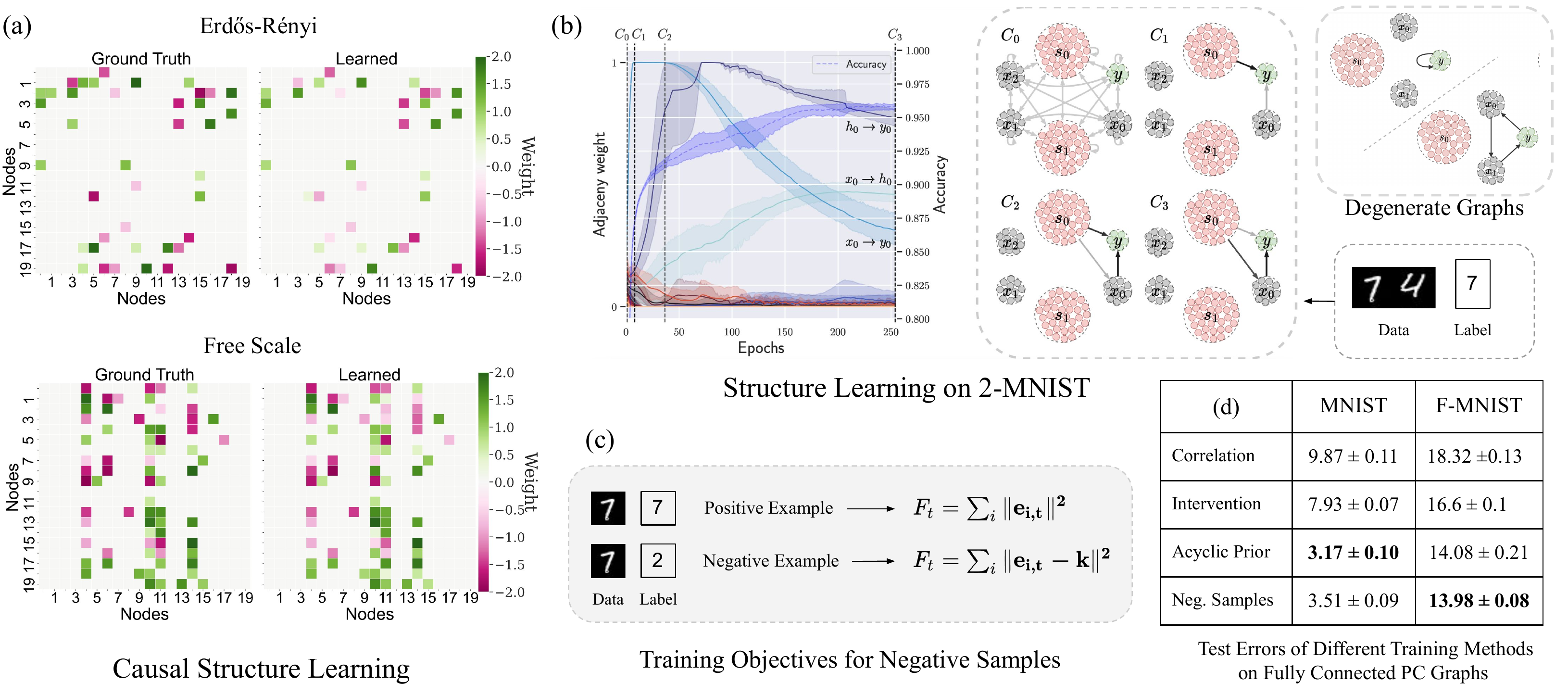}
 \vspace{-4ex}
\caption[short]{(a) Experiments on structure learning from synthetic data, generated from Erd\H{o}s-R\'{e}nyi and scale-free random graphs with $20$ nodes. On the left, the connection strength of the true graph; on the right, the one learned by a PC graph. (b) Structure learning on the $2$-MNIST dataset: the plot shows the weights of the adjacency matrix $\mathbf{A}$ over the number of epochs, the dotted curve  the test accuracy. The vertical lines $C_i$, refer to the connectivities discovered by the structure learning algorithm during training. Such connectivities are shown on the right side of the plot, where you can see the $6$ clusters of neurons, and the connections among them.  For example, the blue one (representing the direct connection), immediately goes to 1, and stays there until the second vertical line (that represents C2), and then starts decreasing. At epoch 250, there are two curves above it: the ones of the hierarchical connections. (c) A description of the two energy functions optimized by the PC graph when training on negative and non-negative examples. (d) Table with test error of all experiments performed on MNIST and FashionMNIST, averaged over three seeds. The best results are obtained when augmenting the training process with both the proposed structure learning methods.}
% https://docs.google.com/drawings/d/1Um_iCdJkY3L3w9ENnBqW9qTcGXvpWdI2bR1qhs4u2DM/edit?usp=sharing

\label{fig:sl_exp}
\vspace{-3ex}
\end{figure*}

We perform two different structure learning experiments. In the first, we are provided with non-interventional data generated by from a Bayesian network of unknown structure. The task is to retrieve the original graph starting with a fully connected PC graph, which is a standard problem in causal discovery \citep{morales2022simultaneous,geffner2022deep,zheng2018dags}. In the second experiment, we perform classification for MNIST and FashionMNIST using a fully connected PC graph. This time, however, we augment the classification objective with priors to enforce sparsity and acyclicity, and conjecture that $(i)$ this improves the final test accuracy, and $(ii)$ reduces a fully connected graph to the ``correct’’ sparse network, i.e., the hierarchical one.
\newpage
%\vspace{-1ex}
\textbf{Structure Learning. \ }%
Here, we use synthetic data sampled from an SCM with $N \,{\in}\, \{10,15,20\}$ nodes and $\{1,2,4\}$ expected edges per node. The graph structure is either an Erd\H{o}s-R\'{e}nyi or a scale-free random graph. To this end, ER2 denotes an Erd\H{o}s-R\'{e}nyi graph with $2N$ expected edges, and SF4 denotes a scale-free graph with $4N$ expected edges. We vary the graph type, number of nodes and/or edges, to test the scalability and stability of each method. We place uniformly random edge weights onto a binary adjacency matrix of a graph, to obtain a weighted adjacency matrix, $\mathbf{W}$. We sample observational data from a set of linear structural equations with additive Gaussian noise.

Due to the linearity, we model each edge as a scalar. Hence, we can set the parameters of the weighted adjacency matrix to be the estimated model parameters $\widehat{\w}$. To prune the parameters not used in the linear structural equations that generate the observed data, we require our model to be sparse and acyclic. Thus, we consider the parameters to have prior distributions $h(\w)$ and $l(\w)$. Then, the experiment consists of training the fully connected model to fit the dataset, and checking whether the PC graph can converge to the random graph structure that generated the data.

The results show that PC graphs are able to infer the structure of the data generating process for arbitrary dense random graphs of various complexities. The heatmaps in Fig.~\ref{fig:sl_exp}(a) show that our method can estimate the true weight adjacency matrix for dense SF4 and ER2 graphs with $20$ nodes. Hence, we conclude that the learned adjacency matrix, which we chose as the median performing model across multiple seeds, is able to well capture the variable dependencies of the observed data. More details on the dataset generation, how the experiment is performed, quantitative results for various structure learning metrics, and detailed comparisons against other baseline methods (PC \citep{kalisch2007estimating}, GES \citep{chickering2002optimal}, NOTEARS \citep{zheng2018dags}, and ICALiNGAM \citep{shimizu2006linear}) are provided in Appendix~\ref{sec:appendix_structure_learning}. The results show that our algorithm sustains a stable performance in all ER and SF graph setups, as validated by structural Hamming distance (SHD), F1 score, and other metrics.

%The curve depicts the graph learning process and the variance of the difference between the true and the estimated structure throughout the learning process. 
%Compared to the baselines  our method performs consistently well across all ER and SF graphs setups, different to other approaches which degrade in performance for some scenarios. Furthermore, Fig.~\ref{fig:SF4_plot} and Fig.~\ref{fig:ER2_plot} show how our PC model is estimating the true weight adjacency matrix for dense SF4 graphs and ER2 plots with 20 nodes.

\textbf{Classification. \ }%
Here, we extend the experiments of Section~\ref{sec:ci}, and check whether the results can be improved by allowing the PC graph to cut extra connections during training. Additionally, we create a new dataset, called $2$-MNIST, whose data consist of pairs $(\s_0,\s_1)$ of MNIST images, and the label is the label of $\s_0$. This is to check whether PC networks are able to understand the underlying causal structure of the dataset, and remove connections that start from $\s_1$. As an architecture, we consider a PC graph with $6$ nodes, one of dimension $784$, one of  dimension $10$, and $4$ hidden nodes of dimension $d$. In the case of $2$-MNIST, we have two nodes of dimension $784$, and only three of dimension $d$. The adjacency matrix $\A$ has then dimension $6 \times 6$. Note that, when the entries of $\A$ are all equal to one, then this model is equivalent to the fully connected one of Section~\ref{sec:ci}. Here, however, we propose two techniques to augment the training process, and let the model converge to a hierarchical network. The first one consists of adding the three proposed priors on the matrix $\A$, to enforce sparsity and acyclicity in the graph; the second one consists of augmenting the dataset via negative examples, while enforcing sparsity via the Laplace prior only. Note that enforcing acyclicity is fundamental, otherwise the circular dependencies in the graph would make the model converge to degenerate structures, such as the ones provided in the top right corner of Fig.~\ref{fig:sl_exp}. More details on this can be found in Appendix~\ref{sec:appendix_classification}.

In the first experiment, we use the $2$-MNIST dataset to test whether the acyclic and sparse priors are able to both remove the out-going connections from the second image $\s_2$, and learn a hierarchical structure, which we know to be the best one to perform classification on MNIST. In the second experiment, we train the same fully-connected model of Section~\ref{sec:ci} and check whether the priors allow to increase the classification accuracy of the model. To conclude, we perform a classification task with the negative examples and the Laplace prior, to test whether this method also allows to avoid converging to degenerated graph structures. 
The results on the $2$-MNIST dataset show that the model immediately prunes the parameters out-going from $\s_2$. In the first $~100$ epochs, the edge with the largest weight is the linear one, which directly connects the input to the label. While this shows that the model correctly learned the dependencies, linear classification on MNIST and FashionMNIST does not yield great accuracies. This problem is naturally addressed in the later stages of the training process, where the entry of the adjacency matrix relative to the linear map loses weights, and hence influence on the final performance of the model. When training finally  converges, the resulting model is hierarchical, with one hidden layer, as shown in the plot in Fig.~\ref{fig:sl_exp}b. This shows that PC graphs are not only able to learn the causal dependencies correctly, but also to be able to discriminate among these structures, and converge to a well performing one.

In the second experiment (classification on MNIST and FashionMNIST with $h(\A)$), the model shows a clear improvement over the baseline in Section~\ref{sec:ci}. The same applies for the training with negative examples (see the table in Fig.~\ref{fig:sl_exp}d), which shows a performance comparable to these of training with an acyclicity prior. To reach the usual results that can be obtained via standard neural networks trained with backpropagation (i.e., a test error $<2\%$), it suffices to fine-tune the model using the newly learned structure.

\vspace{-1ex}
\section{Related Work}
\vspace{-1ex}

In the last years, there have been numerous works that have tackled machine learning problems using PC networks. They have been shown to perform well in classification tasks using all kinds of architectures, such as feedforward and convolutional models, graph neural networks, and transformers \citep{whittington2017approximation,han2018deep, salvatori2022incremental, byiringiro2022robust, pinchetti2022predictive}. These results are partially justified by some similarities that PC shares with backpropagation when performing supervised learning \citep{Song2020,millidge2020predictive,salvatori2021any}. Multiple works have also applied it to  image generation \citep{ororbia2022neural,ororbia2019biologically}, continual learning \citep{ororbia2022lifelong, song2022inferring}, and associative memories \citep{salvatori2021associative, yoo2022bayespcn, tang2023recurrent}. We refer to \citep{salvatori2023braininspired} for a comprehensive review.

Causality has found applications in problems such as treatment effect estimation, time series modeling, image generation, and natural language processing, as well as enhancing interpretability and fairness in machine learning \citep{shalit2017estimating, Runge2019_time, NCC_image, Zachary2020_nlp, kusner2017counterfactual}. Some works study the problem of learning the causal structure from observational data, previously done via combinatorial search \citep{spirtes2000causation,chickering2002optimal,shimizu2006linear,kalisch2007estimating}. However, combinatorial searches algorithm grow double exponentially in complexity relative to the dimension of the graph. To this end, recent works mostly performing continuous optimization by using the acyclic prior that we have also discussed in our work \citep{zheng2018dags,bello2022dagma,yu2019dag}.

\vspace{-1ex}
\section{Conclusion}
\vspace{-1ex}

We have provided a bridge between the fields of causality and computational neuroscience by showing that predictive coding graphs have the ability of both learning the DAG structures from observational data, and modeling associational, interventional, and counterfactual distributions \citep{geffner2022deep, sharma2020dowhy}. This makes our method suitable candidate for an end-to-end causality engine, which can answer causal queries without knowing detailed structural equations of an SCM. In detail, we have shown how interventions can be performed by setting prediction errors of nodes that we are intervening on to zero, and how this leads to the formulation of predictive-coding-based structural causal models. For structure learning, we have shown how to use existing techniques to derive causal relations from observational data. %More generally, this work further highlights the flexibility of predictive coding models, which can be used to both train deep neural networks that perform well on different machine learning tasks, and to perform causal inference on directed graphical models.

\section*{Acknowledgements}
We thank the reviewers for their valuable feedback and insightful discussions, which have significantly enhanced this manuscript. Amine M'Charrak gratefully acknowledges support from the Evangelisches Studienwerk e.V. Villigst through a doctoral fellowship. This work was also supported by  the AXA Research Fund and by the EU TAILOR grant.

\section*{Impact Statement}
In the domain of causal structure learning from observational data, it is crucial to recognize that algorithms can only identify the causal graph within the confines of a Markov Equivalence Class (MEC). This limitation means that various causal graphs, each with differing implications, could be equally supported by the observational data. Our approach does not incorporate interventional data, amplifying the need for caution in its application for causal discovery. The method should be used alongside further assumptions and in consultation with subject-matter experts to minimize the risk of affirming false causal relationships. Incorrect usage has the potential for negative societal consequences, such as poor or biased decision-making processes. 

\bibliography{references}
\bibliographystyle{icml2024}

%%%%%%%%%%%%%%%%%%%%%%%%%%%%%%%%%%%%%%%%%%%%%%%%%%%%%%%%%%%%%%%%%%%%%%%%%%%%%%%
%%%%%%%%%%%%%%%%%%%%%%%%%%%%%%%%%%%%%%%%%%%%%%%%%%%%%%%%%%%%%%%%%%%%%%%%%%%%%%%
% APPENDIX
%%%%%%%%%%%%%%%%%%%%%%%%%%%%%%%%%%%%%%%%%%%%%%%%%%%%%%%%%%%%%%%%%%%%%%%%%%%%%%%
%%%%%%%%%%%%%%%%%%%%%%%%%%%%%%%%%%%%%%%%%%%%%%%%%%%%%%%%%%%%%%%%%%%%%%%%%%%%%%%
\appendix

\newpage

\

\newpage

\section{Learning on PC Graphs}\label{sec:appendix_algo}

Given a labeled point, two phases are needed to perform a single weight update. The first one, called \emph{inference}, is used both in the training phase, to compute the best configuration of value nodes to perform a weight update, and in the prediction phase, to compute an output when provided a specific input. The inference phase corresponds to \emph{Query by conditioning}, as described in Section~\ref{sec:intro}. During this phase,  the weights are frozen, and only the internal value nodes are updated to minimize the free energy. The second phase happens after the inference has converged, and hence the 'best` neural activities are computed. Here, the opposite happens: all the value nodes are now frozen, and a single weight update is performed to further minimize the same energy function. If we are considering models with an adjacency matrix $A$, we also update its parameters. We will now provide a more formal description of the two phases.

Let us assume we are presented with a data point $\mathbf{S}_{data} = \s_{i_1},\dots, \s_{i_n}$. First, the value nodes of the vertices $v_{i_1},\dots, v_{i_n}$ are fixed to be equal to the entries of $\mathbf{S}_{data}$ for the whole duration of the training process, i.e., for every $t$. Second, the variational free energy is minimized  via gradient descent on the value nodes, 
During this phase, the weights are fixed, and the value nodes are updated as follows:
\begin{equation*}
\Delta{\x}_{i,t}   = - \gamma \frac{\partial F_t}{\partial \x_{i,t}} = \gamma\cdot ( -\e_{i,t} + f' ( \x_{i,t} )  {\textstyle\sum}_{k \in ch(i)} \e_{k,t} \w_{i,k}),
\label{eq:x_update}
\end{equation*}
where $\gamma$ is the a positive real number that indicates a learning rate. When the inference phase is completed, the value nodes get fixed, and a single weight update is performed as follows: 
\begin{equation}
\Delta \w_{i,j}  = -\alpha\cdot \frac{\partial F_t}{\partial \w_{i,j}} =  \alpha\cdot \e_{i,T} f ( \x_{j,T}).
\label{eq:theta_update}
\end{equation}
To conclude, the update of the entries of the adjacency matrix (without the priors), are the following:
\begin{align}
 \Delta a_{i,j} =  - \beta \frac{\partial F_t}{\partial a_{i,j}} = \beta\cdot \e_{i,T} \w^{\top} f ( \x_{j,T}).
\label{eq:update_a}
\end{align}
We provide the pseudocode of the training process on PC graphs in Algorithm~\ref{algo:PC_graph}.

\begin{algorithm}[t]
    \caption{Learning a data point $\mathbf{S}_{data} = \s_{i_1},\dots, \s_{i_n}$} \label{algo:PC_graph}
    \begin{algorithmic}[1]
    \REQUIRE  $(\x_{i_1,t},\dots,\x_{i_n,t})$ is fixed to $(\s_{i_1},\dots, \s_{i_n})$ for every $t$.
    \FOR{$t=1$ to $T$}
        \FOR{each vertex $i$}% and each weight $\bm \mathbf{u}ptheta_{i,t}$}
            \STATE update $\x_{i,t}$ to  minimize $F_t$ via Eq.~\eqref{eq:x_update}
        \ENDFOR
        \IF{$t= T$} 
            \STATE update every $\w_{i,j}$ to minimize $F_t$ via Eq.~\eqref{eq:theta_update}.\\
            update every $a_{i,j}$ to minimize $F_t$ via Eq.~\eqref{eq:update_a}.
        \ENDIF
    \ENDFOR
    \end{algorithmic}
\end{algorithm}

%\end{document}
\section{Proof of Theorem 1}\label{sec:appendix_proof}

\begin{proof}[Proof of Theorem 1]
We seek to prove:
\[
E(\x_j \mid do(\x_i = \s)) = E(\x_{j,T} \mid \forall t: \x_{i,t} = \s, \e_{i,t} = 0).
\]

Let $G'$ be the mutilated graph structure of the Bayesian network $G$ after a do-operation. Then, by definition of the expectation of interventional distributions, we have that 
\[
E(\x_j \mid do(\x_i = \s))_{G} = E(\x_j \mid \x_i = \s)_{G'},
\]
where the expectations are computed, respectively, on graph $G$ and $G'$.
We utilize the value node update rule for \( \Delta \x_{i,t} \) defined in Eq. \ref{eq:x_update}. Our aim is to demonstrate that the node values in the PC graphs defined on $G$ (with $\e_i = 0$) and $G'$ follow the same update dynamics and thus have identical distributions.

\textbf{Case 1: Parents of the Intervention Node $x_i$}

In $G$, the value node update rule for any parent $\x_j$ of $x_i$ is:
\[
\Delta \x_{j,t}^{G} = \gamma \cdot (-\e_{j,t} + f'(\x_{j,t}) \sum_{k \in ch(j)} \e_{k,t} \w_{j,k}).
\]
When $\e_i = 0$, the term involving $\e_i$ is omitted, yielding:
\[
\Delta \x_{j,t}^{G} = \gamma \cdot (-\e_{j,t} + f'(\x_{j,t}) \sum_{k \in ch(j) \setminus \{i\}} \e_{k,t} \w_{j,k}).
\]
In $G'$, $\x_i$ is removed due to the do-operation, resulting in an identical update rule:
\[
\Delta \x_{j,t}^{G'} = \Delta \x_{j,t}^{G}.
\]

\textbf{Case 2: The Intervention Node $\x_i$ Itself}

In both $G$ and $G'$, the value $x_i$ remains constant at $\s$, making its update rule irrelevant.

\textbf{Case 3: Children of the Intervention Node $x_i$}

For a child $x_j$ of $x_i$ in $G$, the update rule is:
\[
\Delta \x_{j,t}^{G} = \gamma \cdot (-\e_{j,t} + f'(\x_{j,t}) \sum_{k \in ch(j)} \e_{k,t} \w_{j,k}).
\]
This update rule remains unchanged in \( G' \):
\[
\Delta \x_{j,t}^{G'} = \Delta \x_{j,t}^{G}.
\]

By addressing all three cases, we show that value nodes in $G$ (with $e_i = 0$) and $G'$ follow identical update dynamics. Therefore, the distributions of remaining variables in $G'$ and $G$ (with $\e_i = 0$) are the same, completing the proof.
\end{proof}

\section{Interventional and Counterfactual Inference}\label{sec:appendix_CI}

\begin{figure*}[htbp]
\centering
\includegraphics[width=0.8\linewidth]{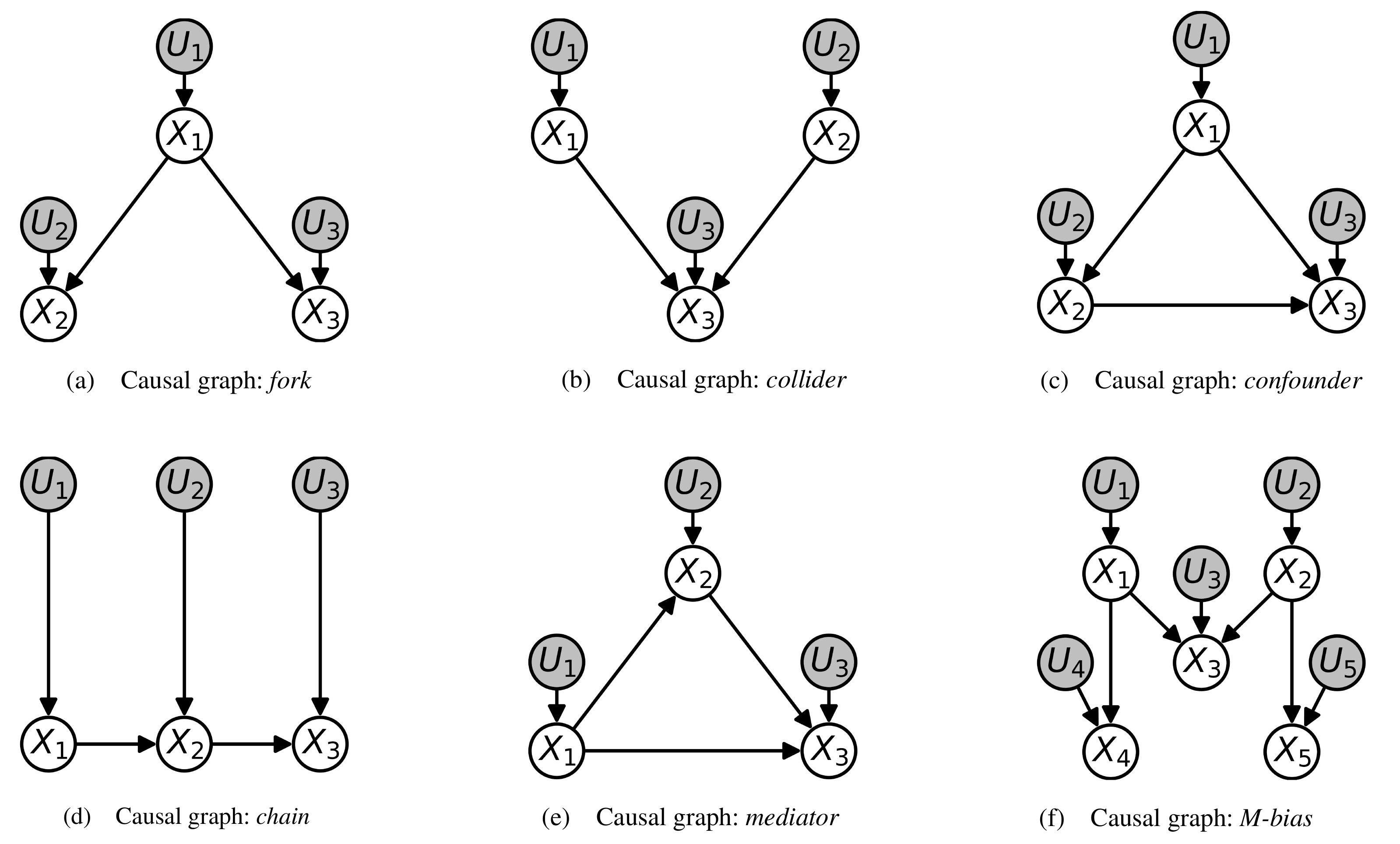}

\caption{Additional graph structures used in experiments of Section~\ref{sec:ci}. White nodes are endogenous variables and observed. Shaded nodes denote independent, exogenous variables for which we do not observe data unless they have no parents in which case $\x_i \coloneqq \mathbf{u}_i$.}
\label{fig:all_DAGs}
\end{figure*}

Here, we provide a detailed discussion on the experiments proposed in Section~\ref{sec:ci}, where we test the ability of PC graphs to model interventional and counterfactual queries. 
The core of decision making is to be able to determine which intervention/action results in an outcome of interest. As such, being able to answer causal queries on a variety of DAG structures and intervention nodes in a DAG is essential.
The causal inference approach that we propose only requires knowledge of the causal structure in the form of parent-child relationships among endogenous variables $\x$. We assume the parameters of the structural equations, $F$, to be unknown.
We make the causal sufficiency assumption, meaning that there is no hidden confounding \citep{peters2017elements}. This is achieved by having an independent exogenous variable for each endogenous variable.

\vspace{-1ex}
\paragraph{Setup.}

We test the associational, interventional, and counterfactual query capabilities of our method on different common directed acyclic graph structures with $N$ endogenous nodes, namely (i) collider ($N = 3$), (ii) confounder ($N = 3$), (iii) mediator ($N = 3$), (iv) chain ($N = 3$), (v) fork ($N = 3$), (vi)~M-bias ($N = 5$), and (vii) butterfly bias ($N = 5$). Each of these structures is visualized in Fig.~\ref{fig:all_DAGs}. For every structure, we generate datasets from a linear SCM with additive Gaussian noise and no restrictions on the location and scale parameters. We use observational training data, $\mathbf{X}$, to fit the PC model. We learn each structural equation, $F_i$, via a training scheme in which we infer the exogenous variables, $\mathbf{u}$, given observed endogenous variables, $\x$, from the training dataset. As such, the SCM is fitted by estimating the parameters of each exogenous variable according to $\mathbf{u}_i \sim \mathcal{N}(\mathbf{u}_i, \sigma_i)$.
This way, we learn to approximate the distribution of the SCM's exogenous variables $\mathbf{u}$.
Note that during the training process, we do not use the factual data once the abduction step is completed. Instead, we replace node values of factual data with inferred values, $\hat{\x}$, by applying the currently learned set of structural equations $\hat{F}$ to the inferred exogenous node values, $\hat{\mathbf{u}}$.
We use associational (\emph{obs}), interventional (\emph{do}), and counterfactual (\emph{cf}) test datasets $\{ \mathbf X^{obs}$, $\mathbf X^{do}$, $\mathbf X^{cf}$ \} to evaluate our method.
The interventions required for $\mathbf X^{do}$ and $\mathbf X^{cf}$ are randomly sampled from $\mathbf{u}_i + \sigma(\x_i) \times \{-1.0, -0.5, -0.1, 0, 0.1, 0.5, 1.0\}$ to ensure realistic intervention values in the support of the each observed marginal distribution.
Here, $\mathbf{u}_i$ and $\sigma(\x_i)$, represent the empirical mean and standard deviation of $\x_i$ from the training data.
To generate an interventional or counterfactual sample, we perform do-operation on individual nodes only, one variable at a time.

For every SCM, we repeat experiments over five different seeds, each using a different PC model initialization. We report error metrics with mean and standard deviation, both multiplied by 100 for clarity.
The PC graph is trained with $3000$ samples for $1000$ epochs with a batch size of $128$. We use the vanilla stochastic gradient descent (\emph{SGD}) optimizer for the node values with a learning rate of $\gamma = 3e-3$ and $T = 8$ iterations for inference of node values during training and testing. For the weights, we use the \emph{AdamW}  optimizer with a learning rate of $\alpha = 8e-3$ and a weight decay of $\lambda_{w} = 1e-4$. 
For linear data, we fit the model using one-dimensional linear layers for each connection between the endogenous and exogenous variables of the SCM according to the causal structure defined given by the adjacency matrix $\A$. In the case of linear data, our approach does not require the use of neural networks with many hidden layers. This makes our causal PC graph transparent, efficient, and lightweight, because we only learn parameters that define the structural equations $F$ of the true data generating SCM.
For the nonlinear experiments, we do not assume any detailed parametric knowledge about the SCMs and our method is feasible with general MLPs as structural equations. Hence, when learning nonlinear data, we replace the linear layers with small MLPs. Each MLP has $2$ hidden layers with $16$ neurons each and we use ELU \citep{ELU_activation} as activation function. 
Note that, despite the large amount of epochs considered, every model converges in less than two minutes. All results are averaged over five random seeds.

\vspace{-1ex}
\paragraph{Metrics.}

We now specify the metrics used to evaluate performance in estimating associational, interventional, and counterfactual distributions, as detailed in Section~\ref{sec:ci}. 
Observational metrics are computed by comparing the true and estimated values of exogenous variables, using the available endogenous node data. 
Furthermore, note that for interventions and counterfactuals only the descendants, $des(\x_i)$, of an intervened node $\x_i$ are affected. Therefore, the causal order of a DAG becomes important when assessing performance on interventional and counterfactual queries. As such, we report metrics with respect to the descendants of the intervention node, as per the adjacency matrix.
We follow the works in the related literature \citep{VACA, chao2023interventional} and report the following metrics:

\begin{itemize}
    \item mean absolute error (MAE), 
    \item maximum mean discrepancy (MMD) \citep{gretton2012kernel}, 
    \item estimation squared error for the mean (MeanE), 
    \item estimation squared error for standard deviation (StdE), 
    \item mean of the squared error (MSE), 
    \item standard deviation of the squared error (SSE).
\end{itemize}

We use MAE as a generic metric to assess the error between the estimated query and the ground truth query. 
The MMD metric is a sample-based distance measure between distributions.
We use MMD to assess the match between the estimated distribution and the true distribution. The idea is to compare the means of both samples, $\widehat{\mathbf{X}}$ and $\mathbf{X}$, in a higher-dimensional feature space defined by a kernel function $k$. 

For a pair of samples from each distribution, we compute the MMD as follows:
\begin{align*}    
& \text{MMD}( \mathbf{X},\widehat{\mathbf{X}}) = \left\lVert \frac{1}{M} \sum_{i=1}^{M} \phi(\mathbf{x}^i) - \frac{1}{M} \sum_{i=1}^{M} \phi(\hat{\mathbf{x}}^i) \right\rVert^2 \\
& = \frac{1}{M^2} \sum_{i=1}^{M} \sum_{j=1}^{M} k(\hat{\mathbf{x}}^i, \hat{\mathbf{x}}^j) - \frac{2}{M^2} \sum_{i=1}^{M} \sum_{j=1}^{M} k(\mathbf{x}^i, \hat{\mathbf{x}}^j) + \\
& + \frac{1}{M^2} \sum_{i=1}^{M} \sum_{j=1}^{M} k(\mathbf{x}^i, \mathbf{x}^j).
\end{align*}

Here, $\phi$ is the feature map of the kernel function $k$, and $\mathbf{x}^i$ and $\hat{\mathbf{x}}^i$ are the $i$-th samples from the ground truth and the inferred data, respectively. 
Each $\hat{\mathbf{x}}^i$ and $\mathbf{x}^i$ is a vector of $N$ features, one for each endogenous node in the DAG. 
The kernel function $k$ measures the similarity between data points in the feature space. In our implementation, we use a mixture of RBF (Gaussian) kernels with varying bandwidth parameters \citep{gretton2012kernel}.

We use MeanE and StdE to assess the estimated interventional distributions. MeanE and StdE measure the average squared error between the true and estimated mean and standard deviation of an interventional distribution, respectively. Both metrics are computed as averages across a set of intervention indices, $\mathcal{I}$, that correspond to nodes in the DAG that have descendants and thus are not leaf nodes. 

Given the empirical means, $E\left[\x_i|do(\x_j)\right]$ and $E\left[\hat{\x}_i|do(\x_j)\right]$, and the empirical standard deviations, $SD\left[\x_i|do(\x_j)\right]$ and $SD\left[\hat{\x}_i|do(\x_j)\right]$, for node, $\x_i$, with intervention on node, $\x_j$, with index $j$, the MeanE and StdE are computed in the following way:
\begin{align}
    &\text{MeanE} = \\
    &\frac{1}{|\mathcal{I}|} 
    \sum\limits_{j \in \mathcal{I}} \frac{1}{|des(j)|} \sum\limits_{i \in des(j)}
    {\left( E\left[\x_i|do(\x_j)\right] - E\left[\hat{\x}_i|do(\x_j)\right] \right)}^2,
\end{align}

\begin{align}
    &\text{StdE} = \\
    &\frac{1}{|\mathcal{I}|}
    \sum\limits_{j \in \mathcal{I}} \frac{1}{|des(j)|} \sum\limits_{i \in des(j)}
    {\left( SD\left[\x_i|do(\x_j)\right] - SD\left[\hat{\x}_i|do(\x_j)\right] \right)}^2.
\end{align}

We denote the number of intervention nodes in the DAG as $|\mathcal{I}|$, and $des(j)$ is the the set of descendants of the intervention node with index $j$. Finally, to assess the performance for the counterfactuals, we report the MSE and SSE for the descendants of an intervention node with index $j$. Both metrics are computed as averages across all intervention nodes in $\mathcal{I}$. We use the \emph{Frobenius norm}, $ T_j = \| \x_{des(j)} - \hat{\x}_{des(j)} \|_F$, to measure the difference between true and estimated values of a counterfactual query with intervention on node index $j$. 
Defining the average of the empirical mean of $T_j$ as $E\left[T_j\right]$ and the average of the empirical standard deviation of $T_j$ as $SD\left[T_j\right]$, we retrieve the MSE and SSE metrics as:
\begin{align}
    \text{MSE} = 
        \sum\limits_{j \in \mathcal{I}} \frac{1}{|des(j)|}
        \,
        E\left[T_j\right], 
\end{align}

\begin{align}
    \text{SSE} = 
        \sum\limits_{j \in \mathcal{I}} \frac{1}{|des(j)|} 
        \,
        SD\left[T_j\right].
\end{align}

To summarize, for associational inference, we report MMD on the observational test set. For interventional inference, we report MMD, MeanE, and StdE. For counterfactual inference, we report MSE and SSE.
Additionally, we report MAE on the associational and interventional inference as well as MSE and SSE for our method's estimates of exogenous noise distributions, which are inferred in the abduction step while performing counterfactual inference.
While not all the above metrics are required to evaluate a model's causal inference performance, we still include them for benchmark comparison against state-of-the-art methods \citep{VACA, CAREFL, MultiCVAE}. 
Across all metrics, lower values indicate better performance.

\section{Experiments on Common Graphs}\label{sec:app_cg}

To generate data from a directed acyclic graph, we first sample a value for each of the $N$ exogenous variables that follow $\mathbf{u}_i \sim \mathcal{N}(\u_i, \sigma_i)$. 
Then, we use the deterministic structural equation, $F_i$, of node $\x_i$ to compute its value as $x_i \coloneqq F_i(par(x_i), \mathbf{u}_i)$. Each $F_i$ is a linear equation with additive noise of the form $F_i = \sum_{j \in par(\x_i)} w_{ji} \x_j + \mathbf{u}_i$, where $par(\x_i)$ denotes the direct parents of node $\x_i$ according to the graph structures provided in Fig.~\ref{fig:all_DAGs}.
We follow the same procedure for the non-linear SCM, however, instead of using linear structural equations, $F_i$, we use non-linear structural equations with additive noise. The non-linear structural equations used to generate the non-linear SCM data are shown in Table~\ref{tab:nonlinear_data_equations}.

\begin{table*}[htbp]
\resizebox{1.0\textwidth}{!}{%
    \centering
    \begin{tabular}{c|cccccc}
        \toprule
        Graph & $F_1 \coloneqq X_1$ & $F_2 \coloneqq X_2$ & $F_3 \coloneqq X_3$ & $F_4 \coloneqq X_4$ & $F_5 \coloneqq X_5$ \\
        \midrule
        \midrule
        Fork & $U_1$ & $-1 + \frac{3}{1+\exp{(-2 X_1})} + U_2$ & $0.25 X_1^2 + U_3$ & - & -\\
        \midrule        
        Collider & $U_1$ & $U_2$ & $0.05X_1 + 0.25X_2^2 + U_3$ & - & -\\
        \midrule
        Confounder & $U_1$ & $-1 + \frac{3}{1+\exp{(-2 X_1})} + U_2$ & $X_1 + 0.25 X_2^2 + U3$ & - & -\\
        \midrule
        Chain & $U_1$ & $-1 + \frac{3}{1+\exp{(-2 X_1})} + U_2$ & $0.25 X_2^2 + U3$  & - & -\\
        \midrule
        Mediator & $U_1$ & $ 1 - \cosh{(0.5 X_1}) + U_2$ & $X_1 + 0.25 X_2^2 + U3$ & - & -\\
        \midrule
        \midrule
        M-bias & $U_1$ & $U_2$ & $0.5 X_1^2 - X_2 + U3$ & $X_1 + 0.5X_1^2 + U4$ & $-1.5 X_2^2 + U_5$\\
        \midrule
        Butterfly & $U_1$ & $U_2$ & $0.5 X_1^2 - X_2 + U3$ & $X_1 + 0.5X_1^2 - 0.25X_3^2 + U4$ & $-1.5 X_2^2 + 0.25X_3^2 + U_5$\\
        \bottomrule
    \end{tabular}%
    }
    
    \medskip % Add space between table and caption
    \caption{Structural equations for non-linear SCM data generation.}
    \label{tab:nonlinear_data_equations}    
\end{table*}

\begin{figure*}[htbp]
    \centering
    \includegraphics[width=0.9\textwidth]{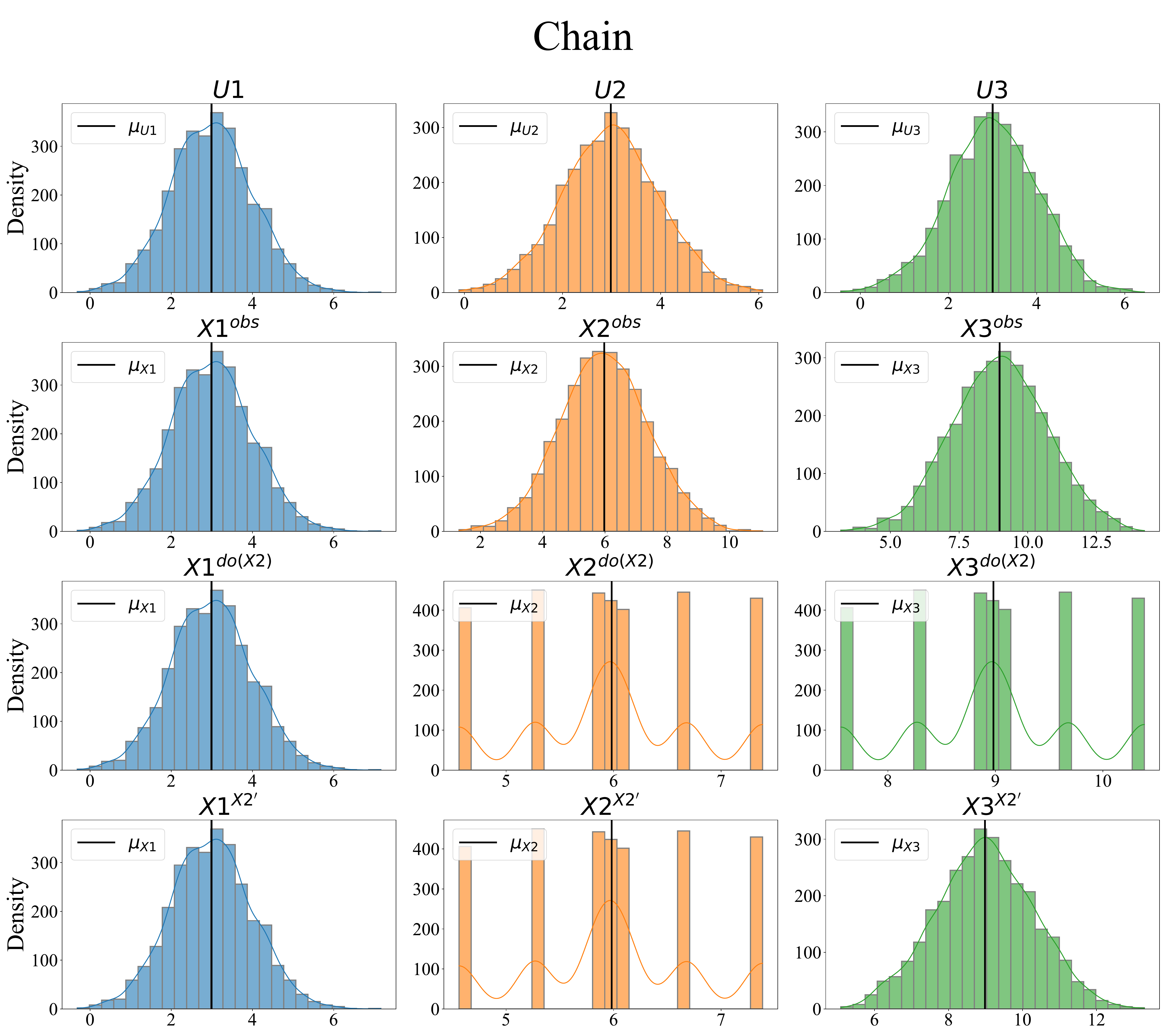}
    \caption{Causal hierarchy of distributions for chain SCM with intervention on $\x_2$. Data generation: linear. First row: Exogenous distribution. Second row: Associational distribution. Third row: Interventional distribution. Last row: Counterfactual distribution.}
    \label{fig:chain_dist}
\end{figure*}

\begin{figure*}[htbp]
    \centering
    \includegraphics[width=0.9\textwidth]{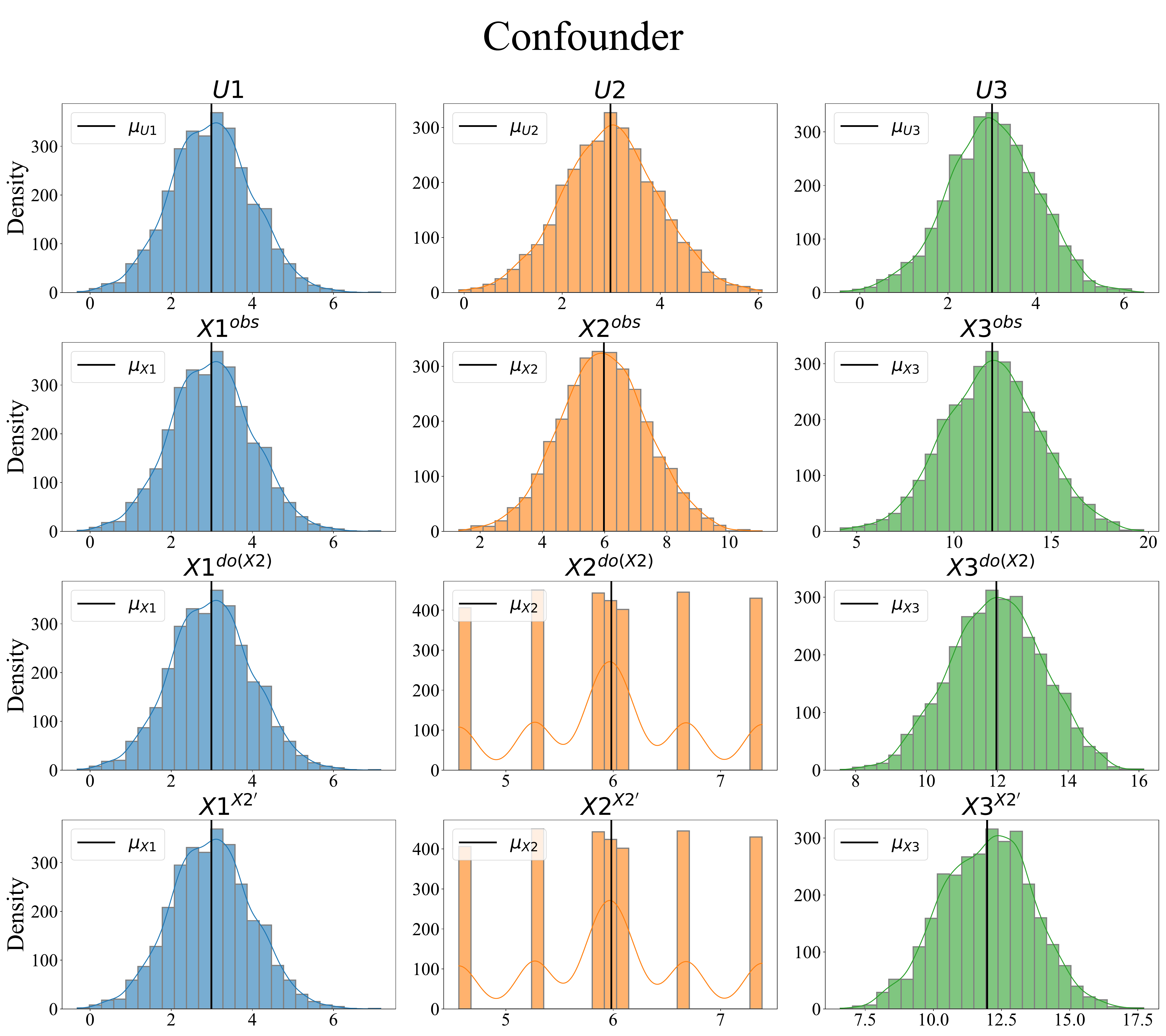}
    \caption{Causal hierarchy of distributions for confounder SCM with intervention on $\x_2$. Data generation: linear. First row: Exogenous distribution. Second row: Associational distribution. Third row: Interventional distribution. Last row: Counterfactual distribution.}
    \label{fig:confounder_dist}
\end{figure*}

\begin{figure*}[htbp]
    \centering
    \includegraphics[width=0.9\textwidth]{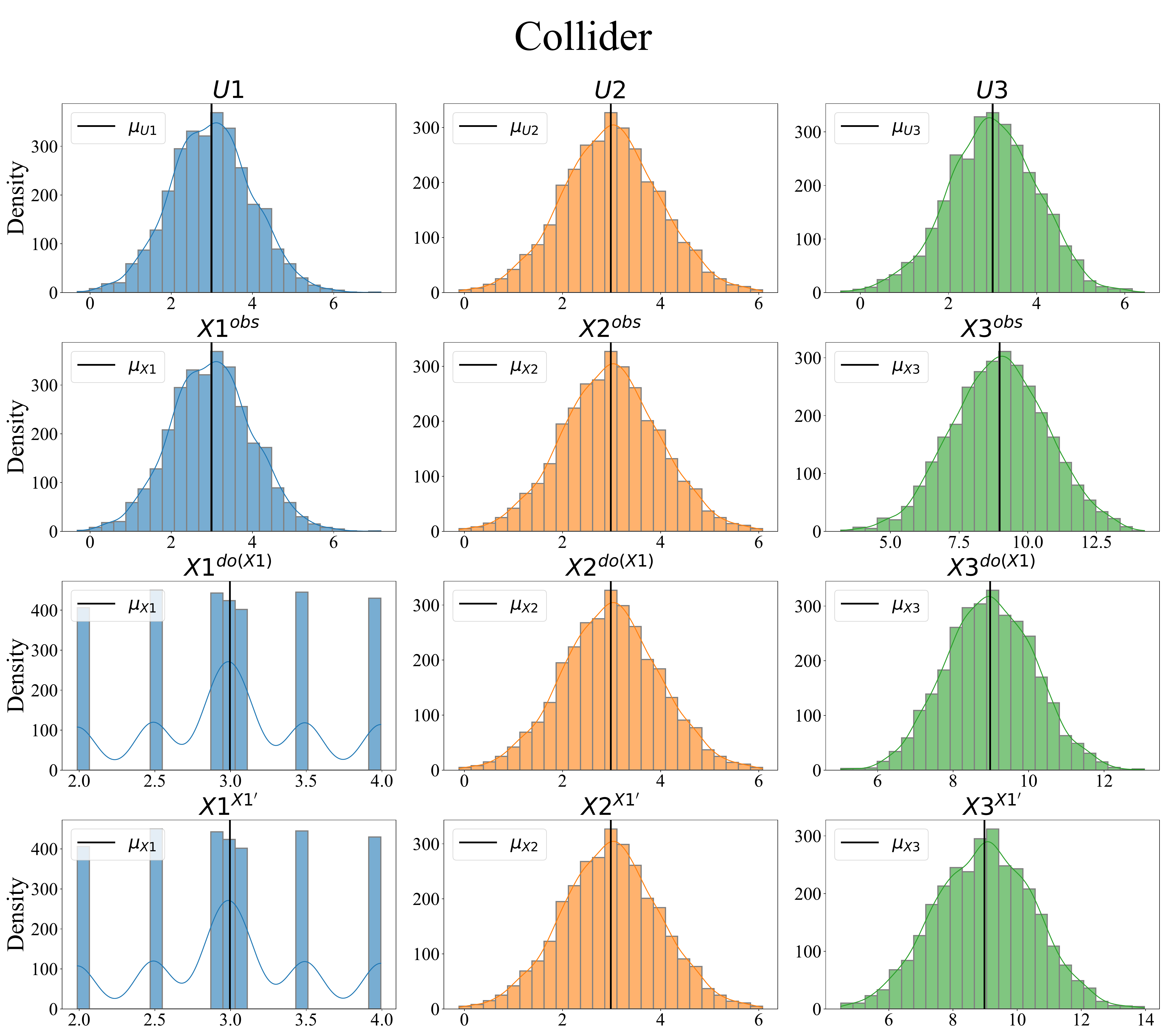}
    \caption{Causal hierarchy of distributions for collider SCM with intervention on $\x_1$. Data generation: linear. First row: Exogenous distribution. Second row: Associational distribution. Third row: Interventional distribution. Last row: Counterfactual distribution.}
    \label{fig:collider_dist}
\end{figure*}

\begin{figure*}[htbp]
    \centering
    \includegraphics[width=0.9\textwidth]{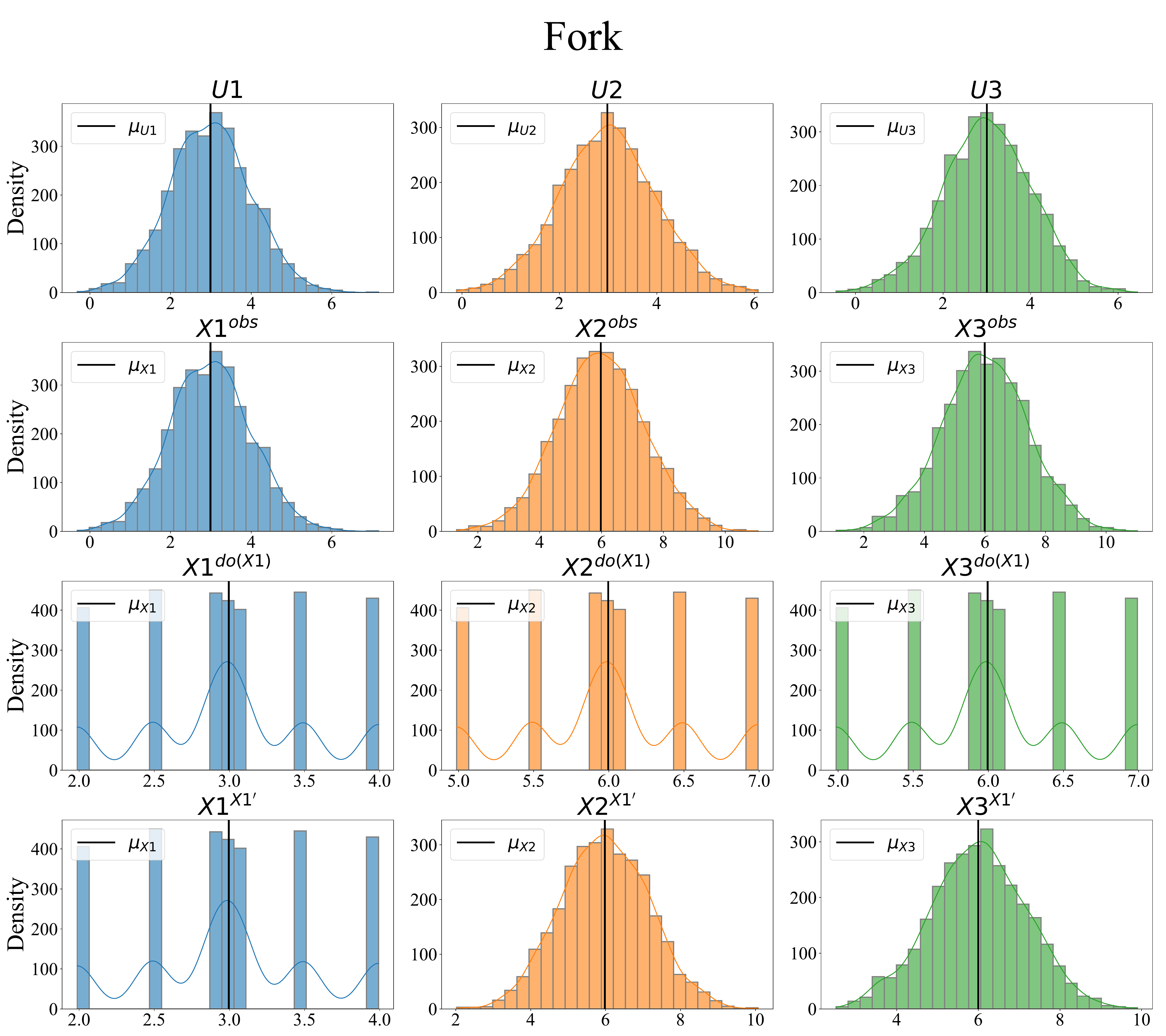}
    \caption{Causal hierarchy of distributions for fork SCM with intervention on $\x_1$. Data generation: linear. First row: Exogenous distribution. Second row: Associational distribution. Third row: Interventional distribution. Last row: Counterfactual distribution.}
    \label{fig:fork_dist}
\end{figure*}

\begin{figure*}[htbp]
    \centering
    \includegraphics[width=0.9\textwidth]{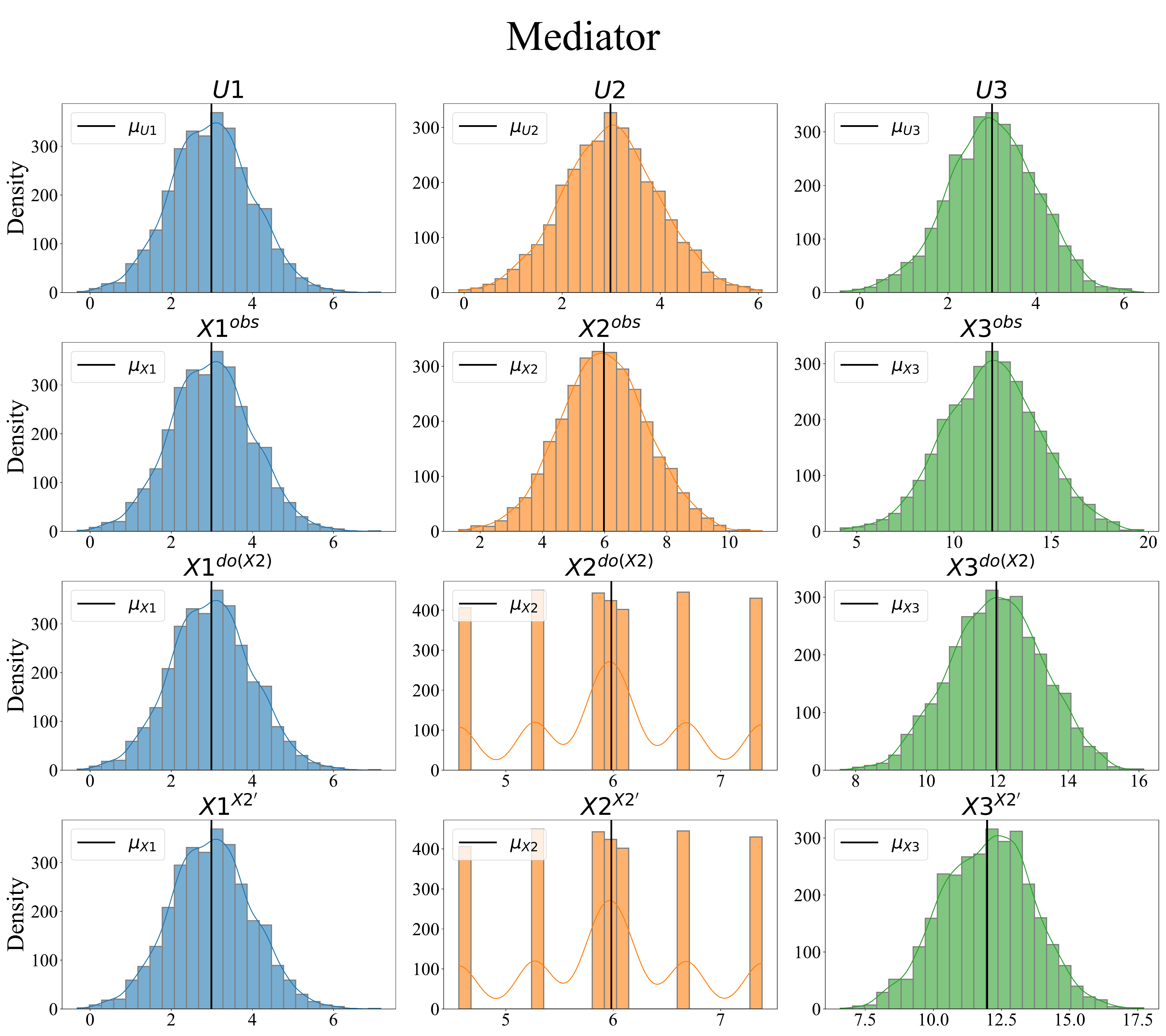}
    \caption{Causal hierarchy of distributions for mediator SCM with intervention on $\x_2$. Data generation: linear. First row: Exogenous distribution. Second row: Associational distribution. Third row: Interventional distribution. Last row: Counterfactual distribution.}
    \label{fig:mediator_dist}
\end{figure*}

\begin{figure*}[htbp]
    \centering
    \makebox[\textwidth][c]{\includegraphics[width=0.99\textwidth]{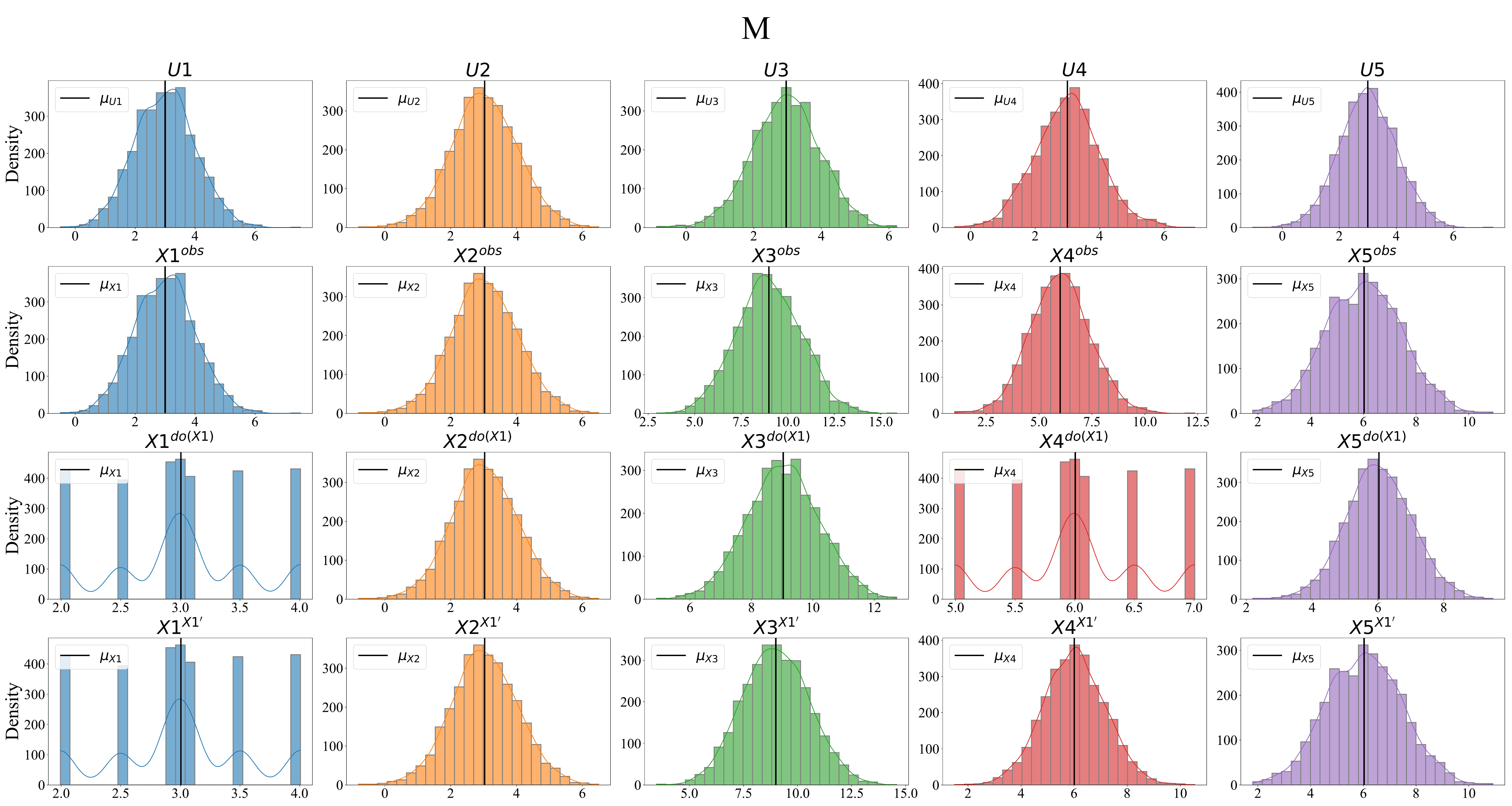}}%    
    \caption{Causal hierarchy of distributions for M-bias SCM with intervention on $\x_1$. Data generation: linear. First row: Exogenous distribution. Second row: Associational distribution. Third row: Interventional distribution. Last row: Counterfactual distribution.}
    \label{fig:Mgraph_dist}
\end{figure*}

\begin{figure*}[htbp]
    \centering
    \makebox[\textwidth][c]{\includegraphics[width=0.99\textwidth]{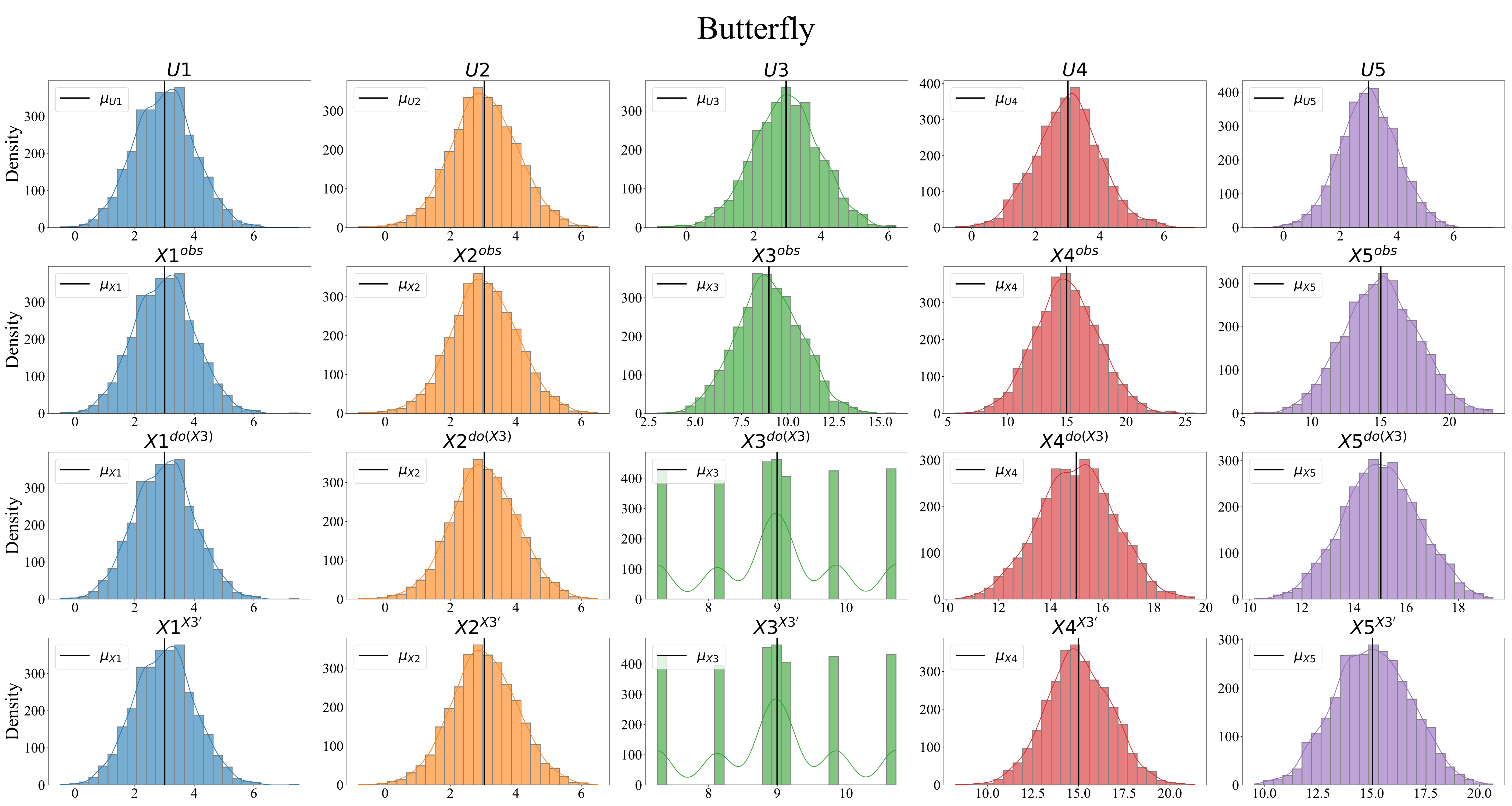}}%    
    \caption{Causal hierarchy of distributions for butterfly SCM with intervention on $\x_3$. Data generation: linear. First row: Exogenous distribution. Second row: Associational distribution. Third row: Interventional distribution. Last row: Counterfactual distribution.}
    \label{fig:butterfly_dist}
\end{figure*}

For the causal inference experiments with the common graphs in Fig.~\ref{fig:all_DAGs} as well as the butterfly graph presented in the main body of the paper, we focused on performing interventions on nodes that are interesting. By interesting we mean that we want to show experimental results for interventions and counterfactuals on nodes that are neither root nodes nor leaf nodes. The reason being that interventions on such nodes either correspond to (a) regular conditional (associational) queries, as is the case with interventions on root nodes or (b) counterfactual queries that are not differentiable from interventional queries, as is the case for interventions on leaf nodes. 
Consequently, we provide results for the following causal inference scenarios: (i) chain graph with intervention on the node $\x_2$, (ii) confounder graph with intervention on node $\x_2$, (iii) collider graph with intervention on root node $\x_1$, (iv) fork graph with intervention on node $\x_1$, (v) mediator graph with intervention on node $\x_2$, (vi) M-bias graph with intervention on node $\x_1$, and (vii) butterfly bias graph with intervention on node $\x_3$. The $\x_3$ intervention in the butterfly graph is interesting and challenging because $\x_3$ is a collider and confounder at the same time. 
Finally, to provide a better understanding of the datasets, what an intervention entails, and how the associational, interventional, and counterfactual distributions differ from each other for each of the graphs depicted in Fig.~\ref{fig:all_DAGs}, we provide the histograms of each \emph{graph-intervention} scenario for the linear data regime in Figs.~\ref{fig:chain_dist} to~\ref{fig:butterfly_dist}. The distribution of each exogenous variable is depicted in the first row as $\mathbf{u}_i$. The histograms show the difference between observational distribution (second row), interventional distribution (third row), and counterfactual distribution (last row), which are denoted as $\x^{obs}$, $\x_i^{do(\x_j)}$, $\x_i^{{\x_i}'}$, respectively (for the purpose of these figures).

\vspace{-1ex}
\paragraph{Results.}

\begin{figure*}[htbp]
\centering
\includegraphics[width=\textwidth, height=0.9\textheight,keepaspectratio]{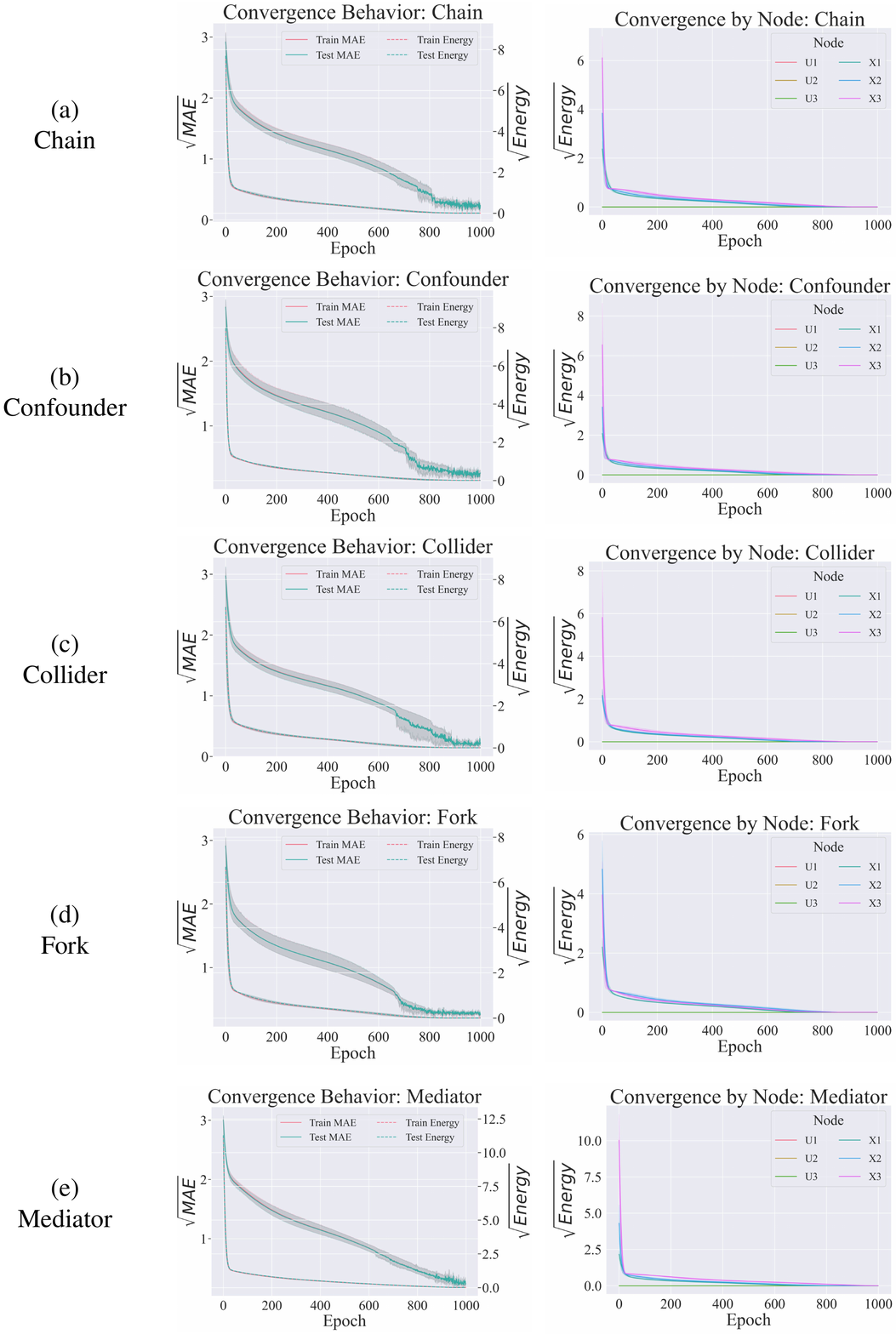}    
\caption{Convergence of energy and MAE by node for causal graphs with five nodes. Left
column: Convergence of total train and test MAE in comparison to free energy. Right column:
Energy by node. Linear data regime. Causal Hierarchy: First level (association query).}
\label{fig:convergence_3_nodes}
\end{figure*}

\begin{figure*}[htbp]
\centering
\includegraphics[width=\textwidth]{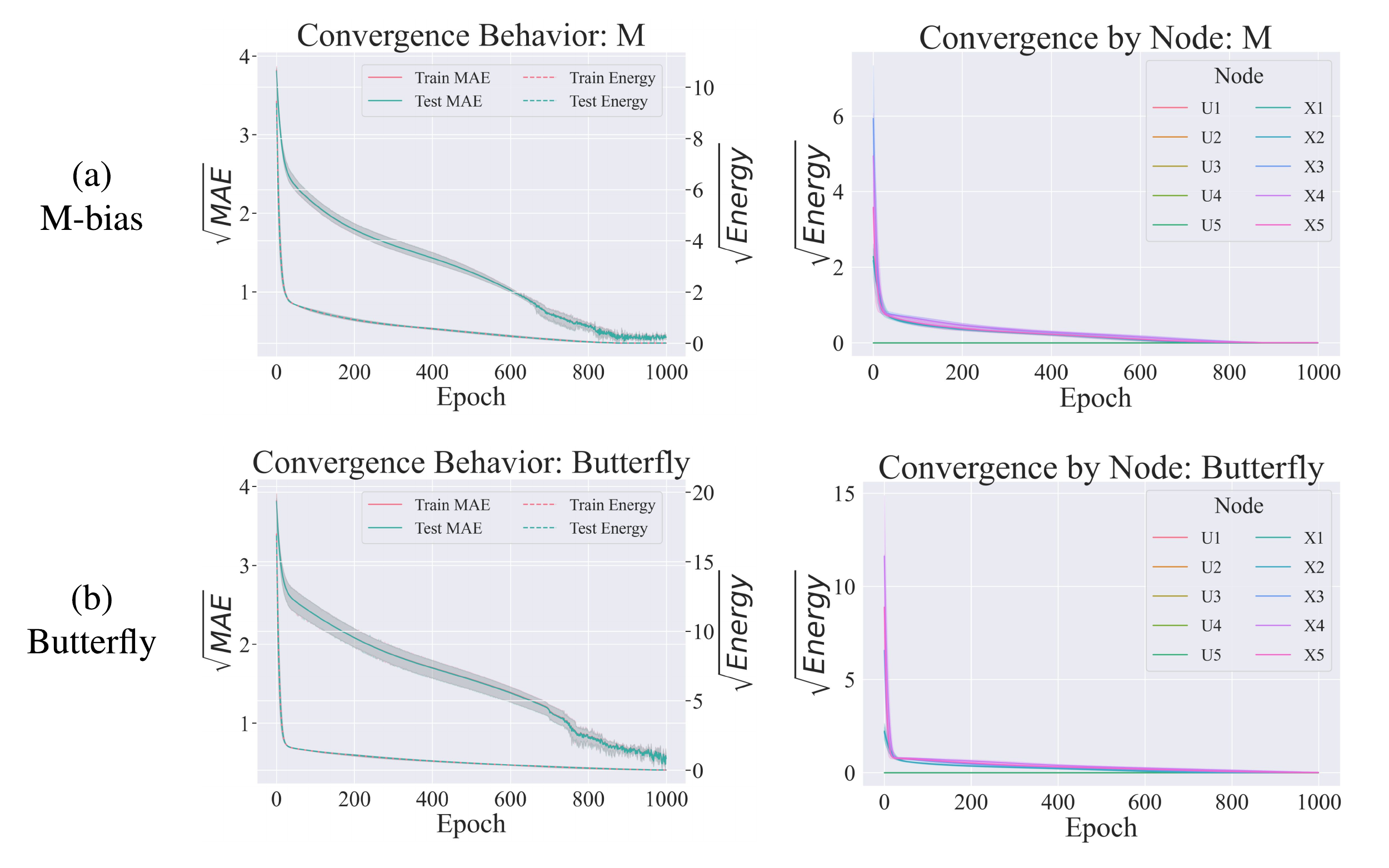}    
\caption{Convergence of energy and MAE by node for causal graphs with three nodes. Left
column: Convergence of total train and test MAE in comparison to free energy. Right column:
Energy by node. Linear data regime. Causal Hierarchy: First level (association query).}
\label{fig:convergence_5_nodes}    
\end{figure*}

\begin{figure*}[htbp]
\centering
\includegraphics[width=\textwidth, height=0.9\textheight,keepaspectratio]{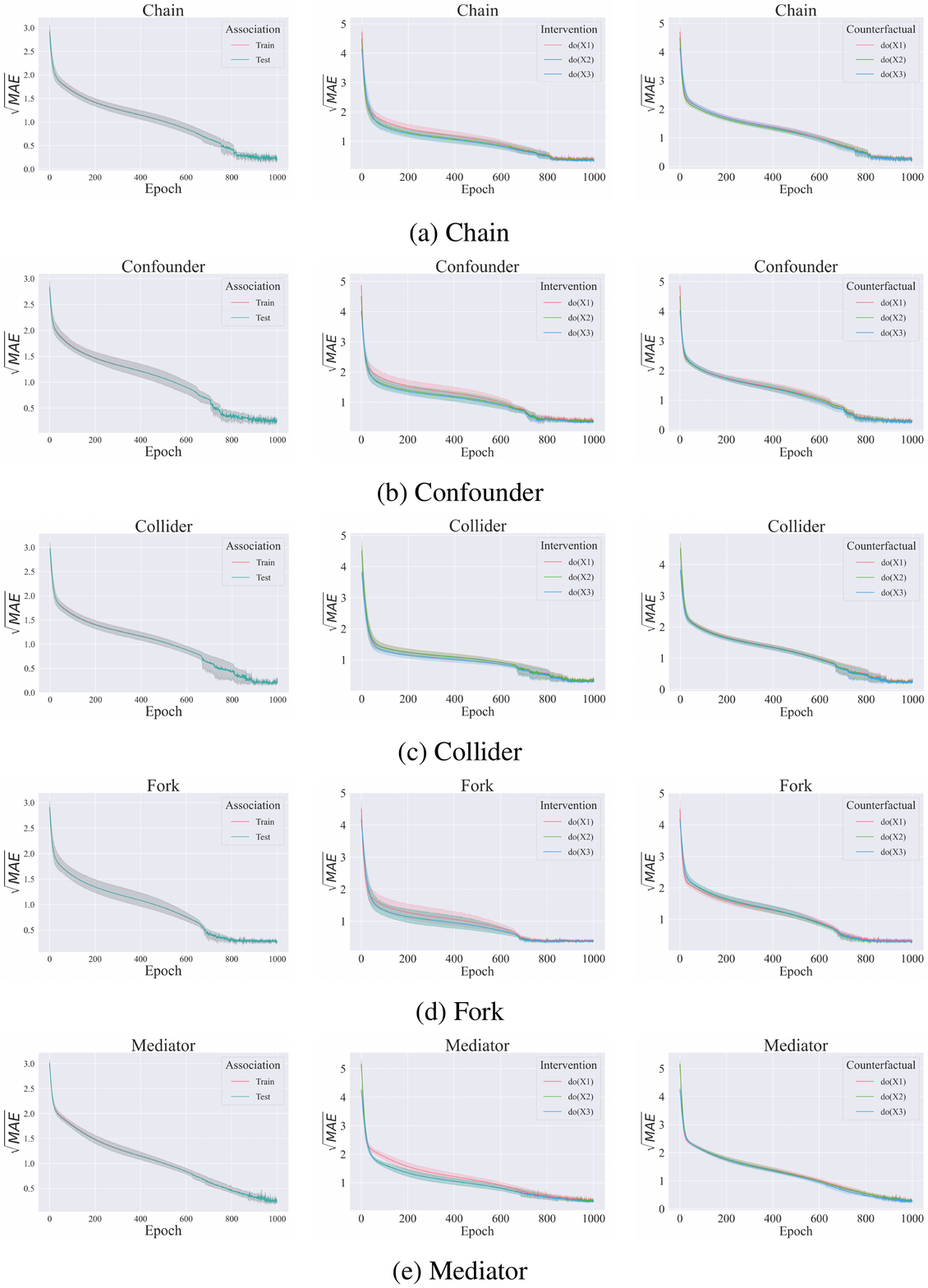}    
\caption{Causal inference performance throughout SCM learning process for linear data regime. We track MAE for all interventions of all three node graphs. For association query (left column), inference is performed on exogenous nodes given factual test data. Center column: Intervention query. Right column: Counterfactual query.}
\label{fig:all_queries_3_nodes}
\end{figure*}

\begin{figure*}[htbp]
\centering
\includegraphics[width=\textwidth]{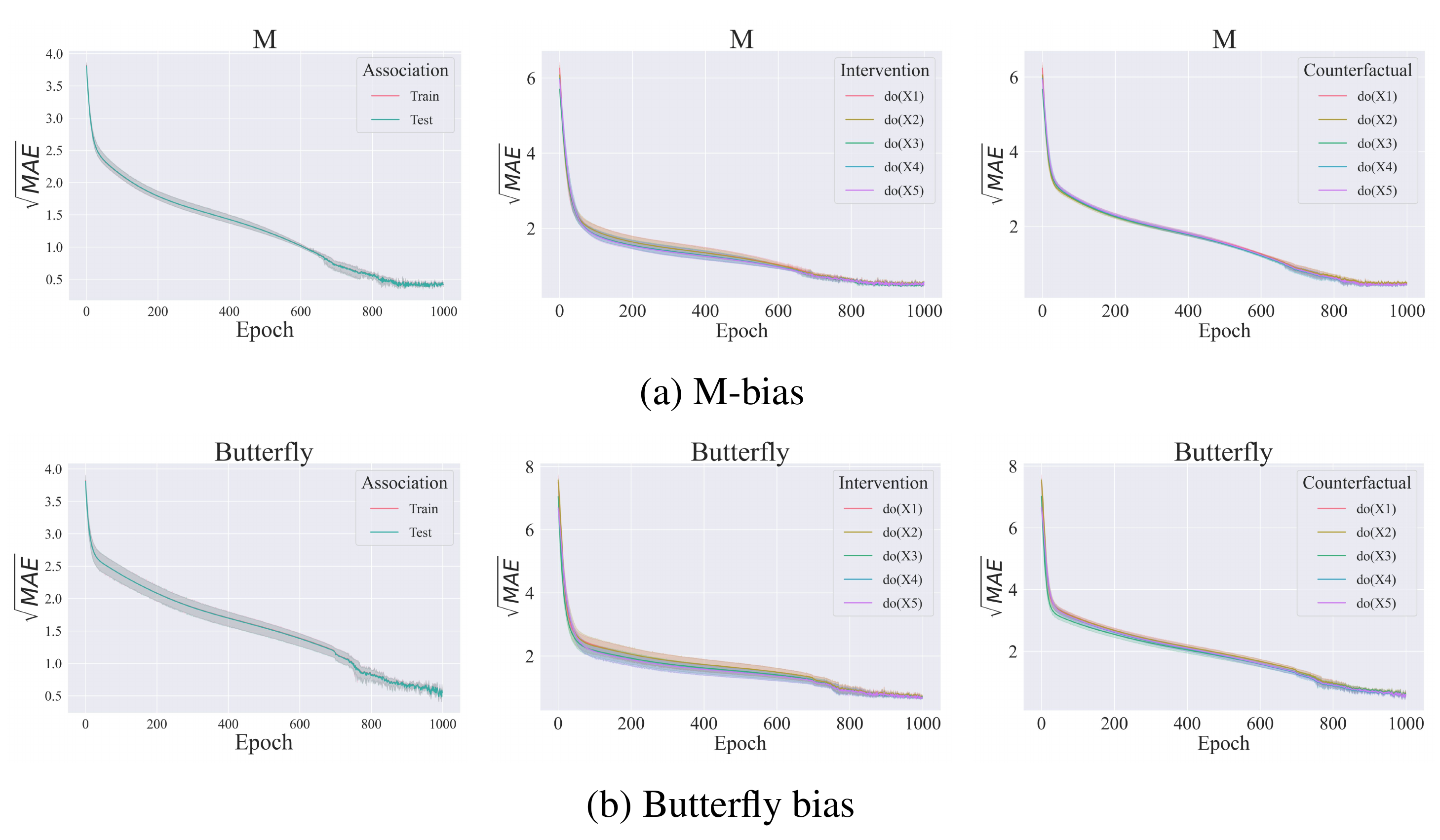}    
\caption{Causal inference performance throughout SCM learning process for linear data regime. We track MAE for all interventions of all five node graphs. For association query (left column), inference is performed on exogenous nodes given factual test data. Center column: Intervention query. Right column: Counterfactual query.}
\label{fig:all_queries_5_nodes}    
\end{figure*}

\begin{figure*}[htbp]
\centering
\includegraphics[width=\textwidth, height=0.9\textheight,keepaspectratio]{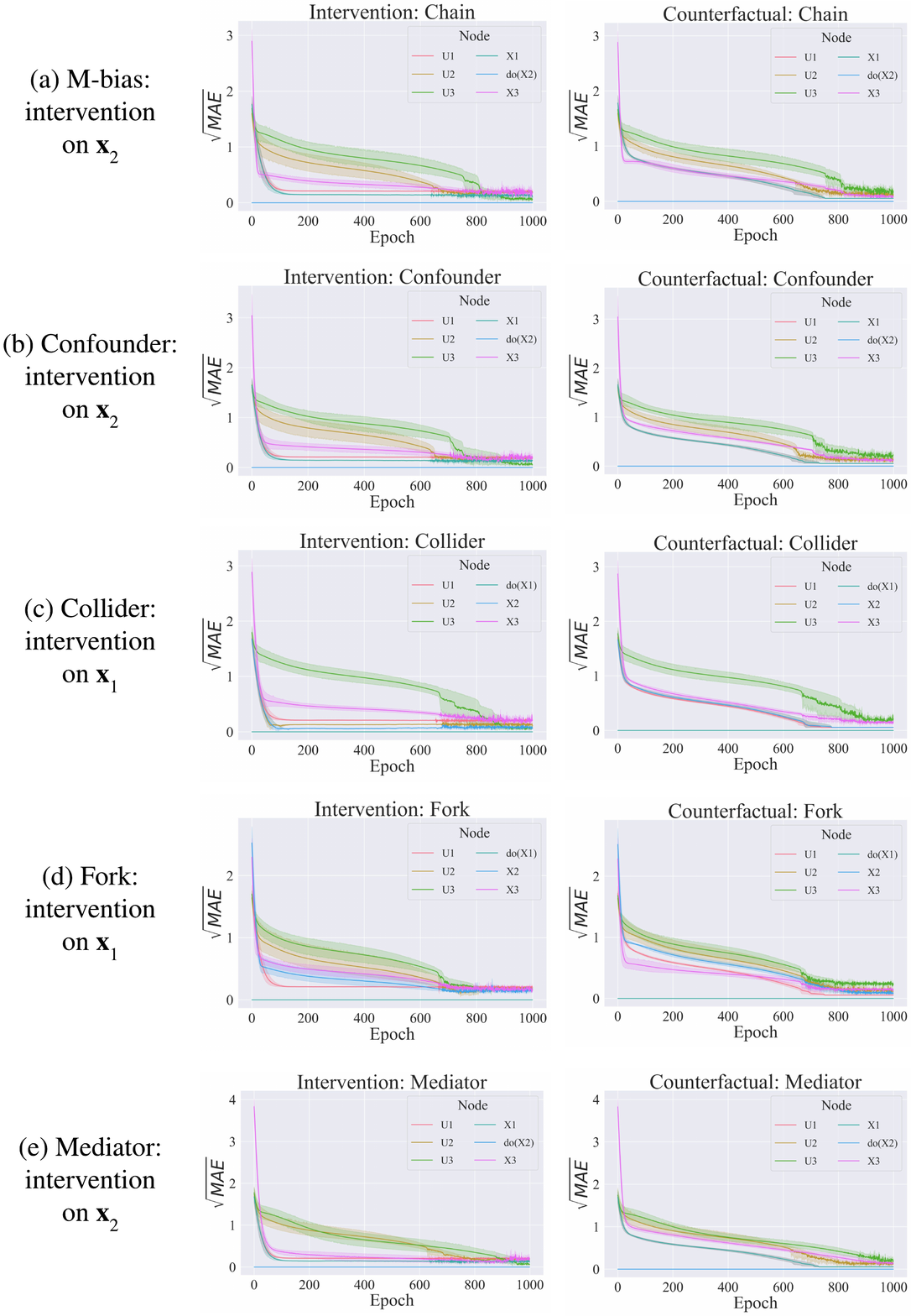}
\caption{Performance of interventional and counterfactual inference throughout SCM learning process for linear data regime. For each three node graph we choose a single intervention node, if available a node that is neither root nor leaf node, and track MAE by node. Left column: Intervention. Right column: Counterfactual.}
\label{fig:unique_intervention_3_nodes}    
\end{figure*}

\begin{figure*}[htbp]
\centering
\includegraphics[width=0.8\textwidth]
{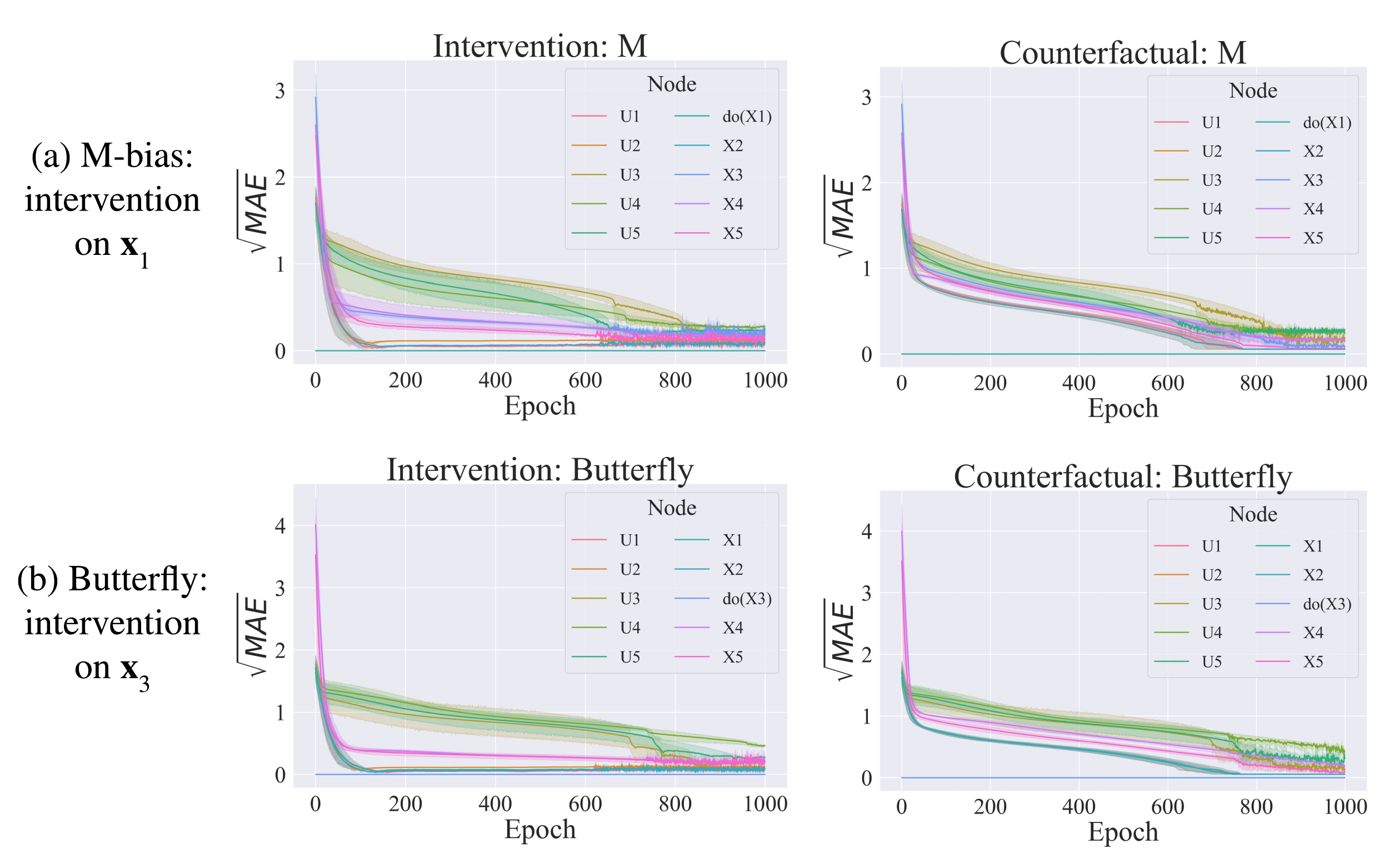}
\caption{Performance of interventional and counterfactual inference throughout SCM learning process for linear data regime. For each five node graph we choose a single intervention node, if available a node that is neither root nor leaf node, and track MAE by node. Left column: Intervention. Right column: Counterfactual.}
\label{fig:unique_intervention_5_nodes}        
\end{figure*}

The experiments in this section display that our method is able to infer correctly associational, interventional, and counterfactual distributions. 
More specifically, we show how we can (1)~learn the parameters of the SCM (structural equations and exogenous distributions) and (2) deploy the error nodes of a fitted PC model in such a way that allows us to manipulate a structural equation to answer causal queries.
First, in Figs.~\ref{fig:convergence_3_nodes} and~\ref{fig:convergence_5_nodes}, we show the convergence of our predictive coding network while learning the parameters of the SCM for all three and five node graphs.
In the left column, we can see that our method converges for all graph structures and that we do not overfit the training data. Moreover, we observed that a low free energy does not correspond to a converged model. The MAE continues to decrease, while the energy changes minimally after 50 epochs.
The right column shows that the convergence is stable and smooth among all nodes in the graph. We show the free energy by node for all exogenous and endogenous variables.
Second, in Figs.~\ref{fig:all_queries_3_nodes} and~\ref{fig:all_queries_5_nodes}, we show the causal inference performance of our method by tracking the MAE for associational, interventional, and counterfactual test queries throughout the SCM learning process. We perform interventions on all types of nodes to show that our model is able to correctly infer causal queries on: root nodes with no parents, intermediate nodes with parents and children, and leaf nodes with no children.
Note that the left column represents the associational inference error on the exogenous nodes only because during training we are provided with data of the endogenous variables.
Third, in Figs.~\ref{fig:unique_intervention_3_nodes} and~\ref{fig:unique_intervention_5_nodes}, we plot the MAE metric for test interventions and counterfactuals during the SCM learning process. We choose intervention nodes that are non-trivial by selecting, wherever available, intervention nodes that are neither root nor leaf nodes.

\vspace{-1ex}
\paragraph{Discussion.}
Our proposed method does not require more parameters than the number of parameters that define the true structural equations of the SCM. 
As such, our model is lightweight and simple to train. 
Having explored various hyperparameters, we found that our model is not prone to overfitting nor does it require hyperparameter tuning or model selection to infer causal distributions.    
%The metrics reported in Table~\ref{tab:SCM_table} show that our method is consistent with the increasing complexity of DAG structures and able to well capture observational, interventional and counterfactual distributions for all graphs.
%(see the metrics for MSE and SSE of the exogenous nodes $U$ during the abduction step). 
Further experiments with varied Gaussian distribution parameters for all exogenous variables confirm the robustness of our causal inference method to these variations. 
%Notably, our approach does not require assumptions like zero mean or unit variance, distinguishing it from methods cited in related works \citep{VACA, saha2022on, chao2023interventional} that do require some of these assumptions for at least one of its exogenous variable. 
%Furthermore, evidenced by the reduced parameter count in Table \ref{tab:SCM_table}, our method is more parameter efficient than the related works \citep{CAREFL, MultiCVAE, VACA}. For linear Structural Causal Models (SCMs), our model is distinguished by its simplicity and efficiency, being the least complex yet adequately explanatory model. Similarly, for non-linear scenarios, our model maintains this efficiency in parameters while outperforming other approaches in all evaluated metrics. Thus, our approach offers an efficient and effective solution for causal inference, that is partially inspired by Occam's razor \citep{OccamsRazor}.
Finally, we do not rely on complex approximators, such as GNN, VAE, gradient boosted regressor or normalizing flow models that require extensive hyperparameter tuning, to learn causal relationships between the observed variables.

%%%%%%%%%%%%%%%%%%%%%%%%%%%%%%%%%%%%%%%%%%%%%%%%%%%%%%%%%%%%%%%%%%%
\newpage

\newpage
\newpage

\newpage
\newpage

\newpage
\newpage

\newpage
\newpage

\newpage
\newpage

\newpage
\newpage

\newpage

\newpage
\newpage

\newpage
\newpage

\newpage
\newpage

\newpage
\newpage

\newpage
\newpage

\newpage
\newpage

\newpage

\section{Classification Experiments}\label{sec:appendix_classification}
Here, we provide all the information needed to reproduce the results of the classification experiments provided in the main body of the paper. Furthermore, we also provide a more detailed study on how the results are affected when changing the parameters of the model.

\vspace{-1ex}
\paragraph{Setup.}
The primary focus of our experiments is on the training of fully-connected neural network models with $2000$ neurons on two datasets: MNIST and FashionMNIST. The chosen architecture for the models is a simple fully connected PC graph. For the models, we conducted an exhaustive hyperparameter search. We opted for a grid search approach, examining several combinations of learning rates for the weights and the latent variables, and subsequently training the models for each combination. The chosen learning rates for the weights were $\{ 0.0001, 0.00005, 0.00001\}$. As for the latent variables, the learning rates tested were $\{1, 0.5\}$, and every batch of $128$ examples was observed for $T \in \{3,5,7\}$ iterations. To optimize the weights of our models, we have used the Adam optimizer; for the value nodes, we have used SGD; as an activation function, ReLU. To conclude, have also tested  \emph{incremental predictive coding} (iPC), a variation of PC that updates the weight parameters at every time step $t$. This method has been shown to improve both the performance and the stability of predictive coding models \citep{salvatori2022incremental}. 
Training was performed for $20$ epochs for each combination of learning rates in the grid search. It is important to note that training always converged before the 20th epoch, ensuring a stable model for each hyperparameter combination. At every epoch of the training process, we have computed the test accuracy using both interventional queries and conditional queries.

\vspace{-1ex}
\paragraph{Results.}
\label{sec:results}

\begin{figure*}
    \centering
    \includegraphics[width=\textwidth]{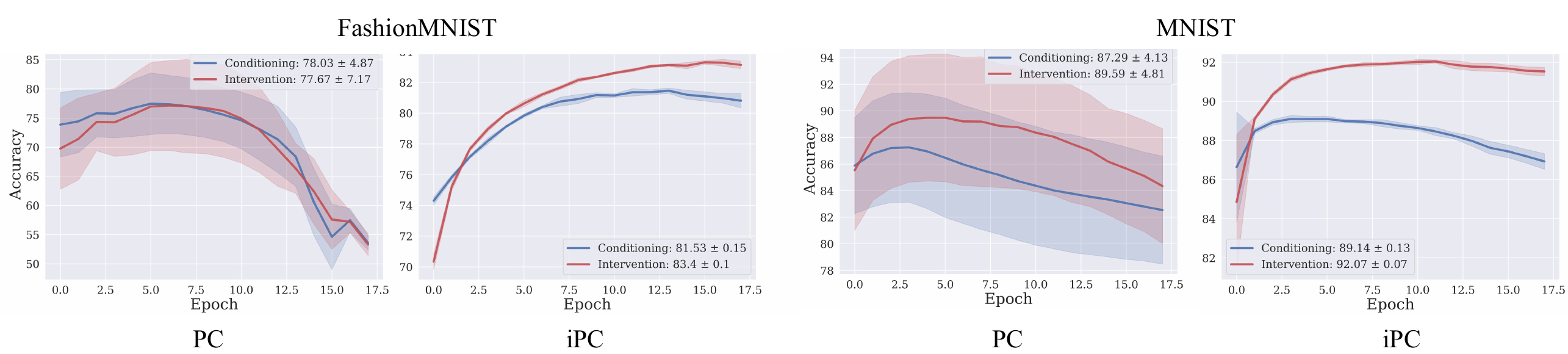}
    \caption{Difference in performance and stability of iPC and PC on classification tasks.}
\label{fig:supp_class_ipc}
\end{figure*}

The results of our experiments showed a clear pattern: regardless of the combination of hyperparameters, interventional queries consistently outperformed conditional queries. This pattern was observed across all models and datasets, suggesting that interventional queries might be a more effective tool for PC graphs. The best results were obtained using a learning rate of the value nodes of $0.5$, and $T = 3$. The learning rate of the parameters slightly affected the performance, unless we consider values outside the proposed range. As a learning algorithm, we observe that iPC is indeed more performing and stable, as shown in Figure~\ref{fig:supp_class_ipc}. 

\newpage

\begin{figure}
    \centering
    \includegraphics[width=0.8\linewidth]{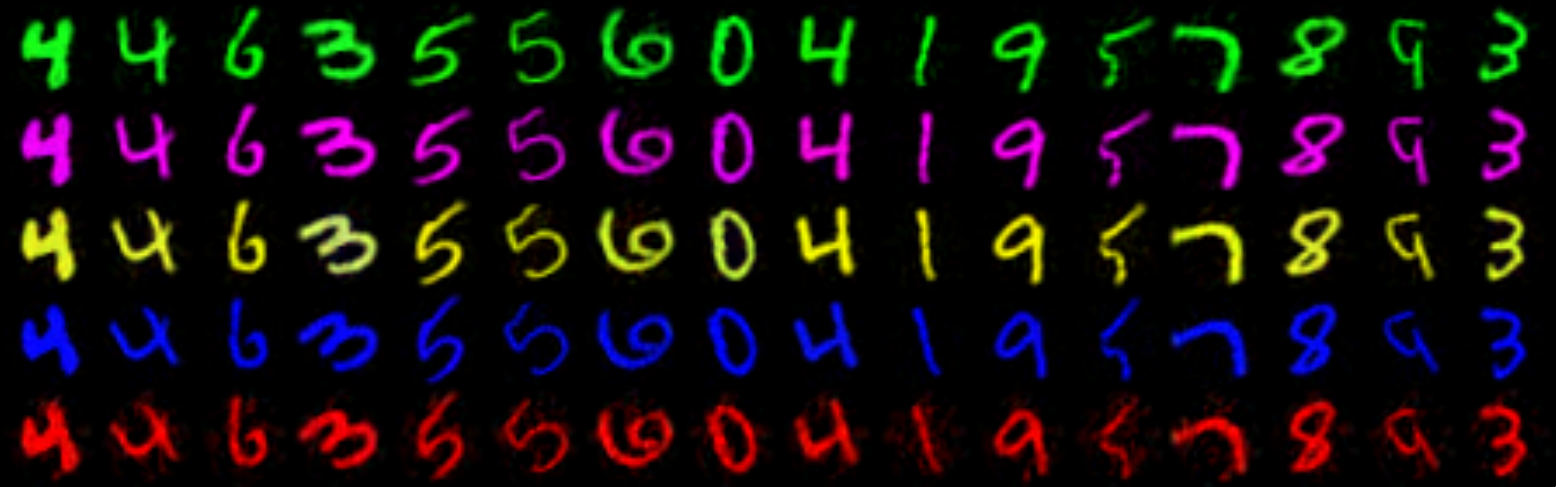}
    \caption{Each line represents digits generated by our predictive coding network when performing two intervention simultaneously (i.e., rotation and color). Rotation angles are, in order, $\mathcal{T} \in \{0, 10, 20, 30, 40\}$. Colors are defined by taking $\mathbb{u}_z$ from the network's state computed for random digits.}
    \label{fig:double_intervention}
\end{figure}

\section{Robustness Experiments}\label{sec:appendix_robustness}

Recent work has shown that available deep-learning-based methods fail to obtain sufficient performance on counterfactual queries under specific circumstances, and proposed novel techniques to overcome this shortcoming \citep{de2022deep}. Here, we show that predictive coding achieves the same state-of-the-art results, while requiring a simpler architecture with no ad hoc training techniques. To do so, we test PC graphs on the colored-MNIST dataset introduced in \citep{de2022deep}. The dataset consists of tuples $(\x, \mathbf{u}_z, \mathcal{T}, \y, \mathcal{T}', \y')$, where $\x$ is the original MNIST image, $\mathcal{T}$ is the assigned treatment, $\mathbf{u}_z$ is a hidden exogenous random variable that determines the color of the observed outcome image $\y$, and $\y'$ is the counterfactual response obtained when applying the alternative treatment $\mathcal{T}'$.
\vspace{-1ex}
\paragraph{Setup.} To replicate the experiment, we have used a PC graph with a structure that is equivalent to the $4$-nodes SCM used in the original work, and trained it with Algorithm~\ref{algo:PC_graph}. We used nodes with the ground truth number of dimensions (i.e., $784$, $1$, and $784$ respectively) to represent the observed variables $\x$, $\mathcal{T}$, and $\y$. Instead, we left the dimension of the remaining hidden node $h$ as a hyperparameter $d_{h}$ as the value $\mathbf{u}_z$ is never observed by the model. The edges between the nodes in the PC graph represent feedforward networks. Figure~\ref{fig:caus_exp}(c) summarizes the architecture used. We experimented with different depths, widths, and activation functions, without experiencing any unexpected results (e.g., deeper and wider networks would have slightly better performance). The architecture can be seen as an encoder-decoder structure. The nodes $\x$, $\mathcal{T}$, and $\mathbf{u}_{z}$ are encoded using respectively $3$, $1$, and $1$ fully connected layers with a hidden dimensions of $1024$ and \emph{tanh} as activation function. Then, the embedding is decoded into $\y$ using $3$ other fully connected layers with the same hidden dimension of $1024$. The experiment was conducted as follows:
\begin{itemize}
    \item During training, we fix the nodes $\x$, $\mathcal{T}$, and $\y$ to the corresponding observed variables and we initialize $h$ to $0$. We train for $128$ epochs with a batch size of $256$. We train using \textit{iPC} and $T=16$. We set the nodes learning rate to $\gamma = 0.005$ and the weights learning rate to $\alpha = 0.00005$. We use the \textit{SGD} optimizer for the nodes and the \textit{Adamw} optimizer for the weights. Consequently, the model has never direct access to any of $\mathbf{u}_z$, $\mathcal{T}'$, or $\y'$. 
    \item The inference process is divided in two phases. Firstly, we repeat the same procedure as above, while setting $\alpha = 0.0$, so that the weights of the model are not changed. This allows the network to adapt to the provided $\y$ by storing its extra information (i.e., the color, in this instance) in the hidden node $\mathbf{u}_z$. Secondly, for each sample in a batch, after $T=16$ steps, we replace $\mathcal{T}$ with $\mathcal{T}'$ to compute the counterfactual $\y'$ and compare it with the ground truth image. To obtain $\y'$, we simply forward through the network the information stored in the nodes $\x$, $\mathcal{T}$, and $\mathbf{u}_z$ during the first phase. To produce the digits in Fig.~\ref{fig:caus_exp}(c), we fixed the node $\mathcal{T}$ to each angle $\in \{0^\circ, 10^\circ, 20^\circ, 30^\circ, 40^\circ\}$. Furthermore, in the main body we show that our model is able to produce counterfactual not only by modifying the rotation $\mathcal{T}$, but also the color encoded by $\mathbf{u}_z$. To show this, we take the value of the node $\mathbf{u}_z$ computed for the a sample and use it to generate all the remaining $\y'$ in the batch. In Figure \ref{fig:double_intervention}, we show that our model is robust also to two complementary interventions, by specifying both rotation and color (acting on variables $\mathcal{T}$ and $\mathbf{u}_z$ respectively) for each digit in the sequence. 
\end{itemize}

\vspace{-1ex}
\paragraph{Results.} We obtain results comparable with the ones in \citep{de2022deep}. Our method has the advantage of using a straightforward multi-layer perceptron architecture trained with an unmodified version of the predictive coding learning algorithm. This shows the capability and versatility of predictive coding to work in various tasks in which other deep-learning techniques tend to fail, such as Diff-SCM \citep{sanchez2022diffusion}, Deep-SCM \citep{pawlowski2020deep}, and Deep-ITE \citep{shalit2017estimating}. Figure~\ref{fig:caus_exp}(c) in the main body shows a magnified example of counterfactual reconstructions that demonstrate that our method is robust with respect to interventions on either rotation or color. Compared to \citep{de2022deep}, we are able to generalize to rotations of $40^\circ$, even if this introduces some noise in the generated output. Furthermore, contrary to the model presented in \citep{de2022deep}, our architecture is robust with respect to the choice of the hyperparameter linked to $u_z$ and does not necessitate to perform a hyperparameter sweep to find the right value.

\newpage

\vspace{-1ex}
\section{Structure Learning}\label{sec:appendix_structure_learning}
\vspace{-1ex}

\vspace{-1ex}
\subsection{Experiments on Random Graphs}
\vspace{-1ex}

In this section, we show results on the convergence behavior of structure learning with a PC graph to understand the relationship between variational free energy and the approximation error of the weighted adjacency matrix. Furthermore, we describe in detail  the metrics used to evaluate the estimated weighted adjacency matrix as well as accuracy metrics for assessing the learned relationships and directions of the adjacency matrix. We also provide all details on the model and training parameters used to reproduce our structure learning experiments.
Finally, we compare our method against established structure learning algorithms for random graphs of various types and complexities.

\vspace{-1ex}
\paragraph{Setup.}

Our causal structure learning method only requires observational data as input. The two types of random graphs that we consider for our experiments are (1) Erd\H{o}s-R\'{e}nyi (ER) with either 1 or 2 expected edges per node, denoted as ER1 and ER2, and (2) scale-free (SF) graphs with either 2 or 4 expected edges per node, denoted as SF2 and SF4, respectively. We use graphs with $N \in \{10,15,20\}$ nodes and generate datasets with $2000$ samples. These two graph types are selected to demonstrate the versatility and robustness of our method in handling various graph structures.

We generate synthetic data by first sampling a binary adjacency matrix, $\mathbf{A}$, for a DAG. Next, we place uniformly random edge weights onto the binary adjacency matrix, to obtain a weighted adjacency matrix, $\mathbf{W}$. Finally, we sample observational data based on a set of linear structural equations with additive Gaussian noise, $\mathbf{u} \sim \mathcal{N}(\mathbf{0}, \mathbf{I}_N)$, such that 
\begin{equation*}
    \x = \mathbf{W}^T \x + \mathbf{u} \in \mathbb{R}^N.
\end{equation*}

To fit the PC model to the observational data, we use the stochastic gradient descent (\emph{SGD}) optimizer for the node values with a learning rate of $\gamma = 1e-4$ and $T = 16$. For the weights, we use the \emph{Adamw} optimizer with a learning rate of $\alpha = 5e-3$. We enforce two penalties onto our learning algorithm. First, a DAG penalty to ensure that the discovered graph is acyclic and directed, as proposed in \citep{zheng2018dags}. 
Second, an L1 penalty that encourages the PC network to find a causal structure that is sparse. We add both penalties into the predictive coding objective. The penalties are each weighted by $\lambda_{L1} = 5e-6$ and $\lambda_{DAG} = 200$, for L1  and DAG penalty, respectively.

\vspace{-1ex}
\paragraph{Structure learning metrics}

Here, we describe the metrics used to evaluate the performance for estimating graph structures in the experiments of Section~\ref{sec:sl}. 
First, for the weighted adjacency matrix of a DAG with $N$ nodes, we report the mean absolute error (MAE) between the true, $\mathbf{W}$, and the estimated, $\widehat{\mathbf{W}}$, weighted adjacency matrix as the average of the absolute differences between corresponding entries in the two matrices:
\begin{equation}
\text{MAE}(\mathbf{W}, \widehat{\mathbf{W}}) = \frac{1}{N^2} \sum_{i=1}^{N} \sum_{j=1}^{N} |\mathbf{W}_{ij} - \widehat{\mathbf{W}}_{ij}|.
\label{eq:sl_MAE}
\end{equation}

Second, to evaluate the correctness of the learned edge directions and the adjacency relationships in the graph, we report metrics on the estimated binary adjacency matrix, $\widehat{\mathbf{A}}$, that is obtained via thresholding as follows:
\begin{equation*}
    \widehat{\mathbf{A}}_{ij} = 
        \begin{cases} 
            1 & \text{if } \widehat{\mathbf{W}}_{ij} > \omega \\
            0 & \text{otherwise.}
        \end{cases}
\label{eq:W_to_A_thresh}
\end{equation*}

We use the same data as proposed in \citep{zheng2018dags}, and we follow their procedure and use $\omega = 0.3$ in all experiments. We report the following graph metrics: (i) F-score (F1), (ii) structural Hamming distance (SHD), (iii) false discovery rate (FDR), (iv) true positive rate (TPR), (iv) false positive rate (FPR), and (v) number of directed edges discovered (NNZ). Each metric is computed between the true adjacency matrix, $\mathbf{A}$, and the estimated adjacency matrix, $\widehat{\mathbf{A}}$. To compute each metric, we first need the following quantities:

\begin{itemize}
    \item true positive (TP): a discovered edge, with correct direction,
    \item reverse (R): a discovered edge, with incorrect direction,
    \item false positive (FP): a discovered edge, not present in $\mathbf{A}$,
    \item true negative (TN): a non-discovered edge, not present in $\mathbf{A}$,
    \item false negative (FN): a non-discovered edge, present in $\mathbf{A}$,
    \item missing (M): a non-discovered edge, present in $\mathbf{A}$.
    
\end{itemize}

Based on these quantities, each metric is computed as follows:

\begin{itemize}
    \item $\text{F1} = \frac{2\text{TP}}{2\text{TP}+\text{FP}+\text{FN}}$, %done
    \item $\text{SHD} = \text{R} + \text{M} + \text{FP}  = \sum_{i=1}^{N} \sum_{j=1}^{N} |\mathbf{A}_{ij} - \widehat{\mathbf{A}}_{ij}|$, % FP == predicted edges that are halucinated (i.e. extra edges)
    \item $\text{FDR} = \frac{\text{FP}+\text{R}}{\text{FP}+\text{TP}}$, % /predicted positives
    \item $\text{FPR} = \frac{\text{FP}+\text{R}}{\text{FP}+\text{TN}}$, % /negatives
    \item $\text{TPR} = \frac{\text{TP}}{\text{TP}+\text{FN}}$, % /positives
    \item $\text{NNZ} = \text{TP} + \text{FP}$.
\end{itemize}

\vspace{-1ex}
\paragraph{Results.}

\begin{figure*}[htbp]
\centering
\includegraphics[width=0.9\textwidth]{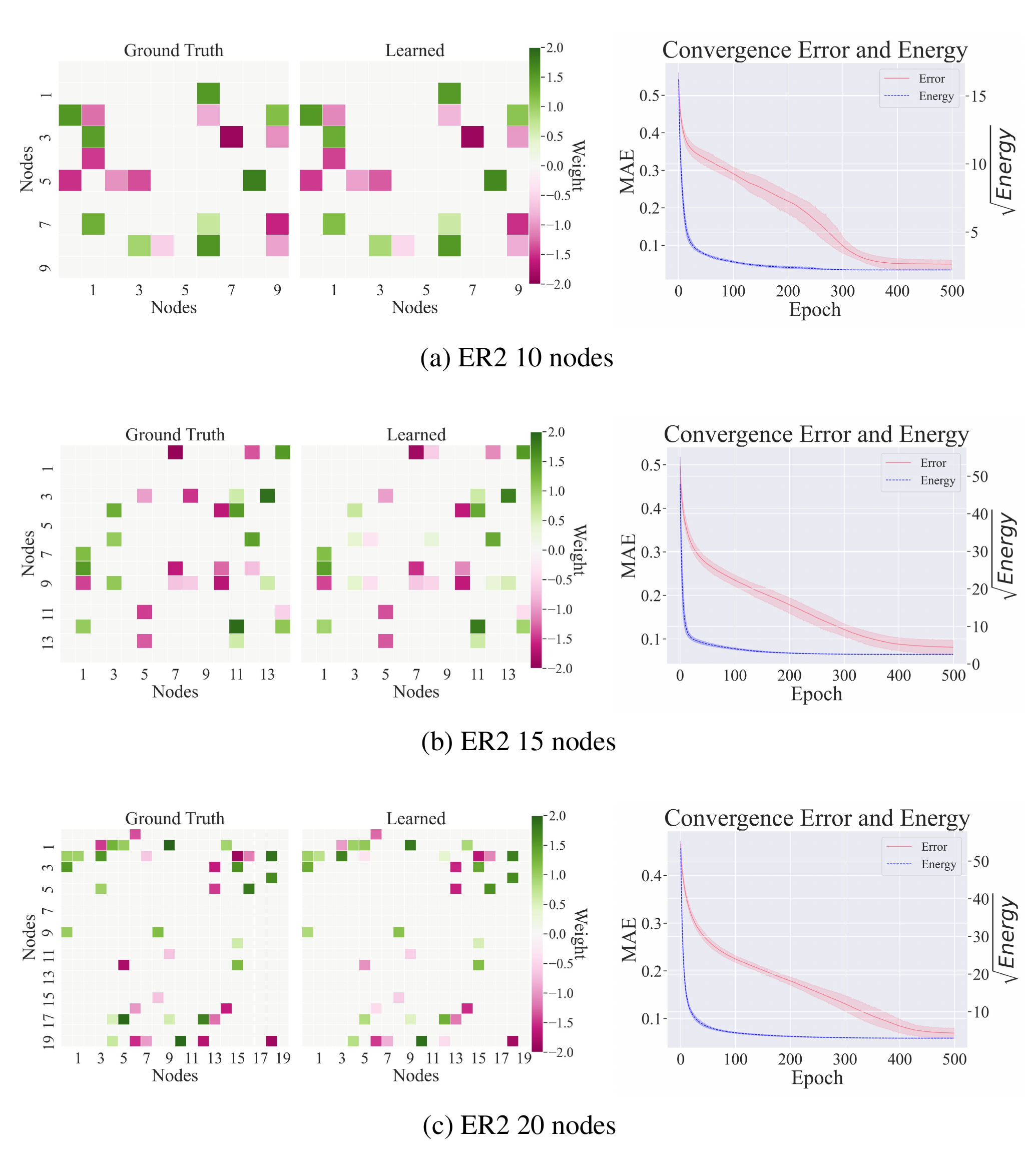}
\caption{Learned structures and convergence behavior (energy vs.\ MAE) for ER2 graphs of various complexity.}
\label{fig:ER2_cl}
\end{figure*}

\begin{figure*}[htbp]
\centering
\includegraphics[width=0.9\textwidth]{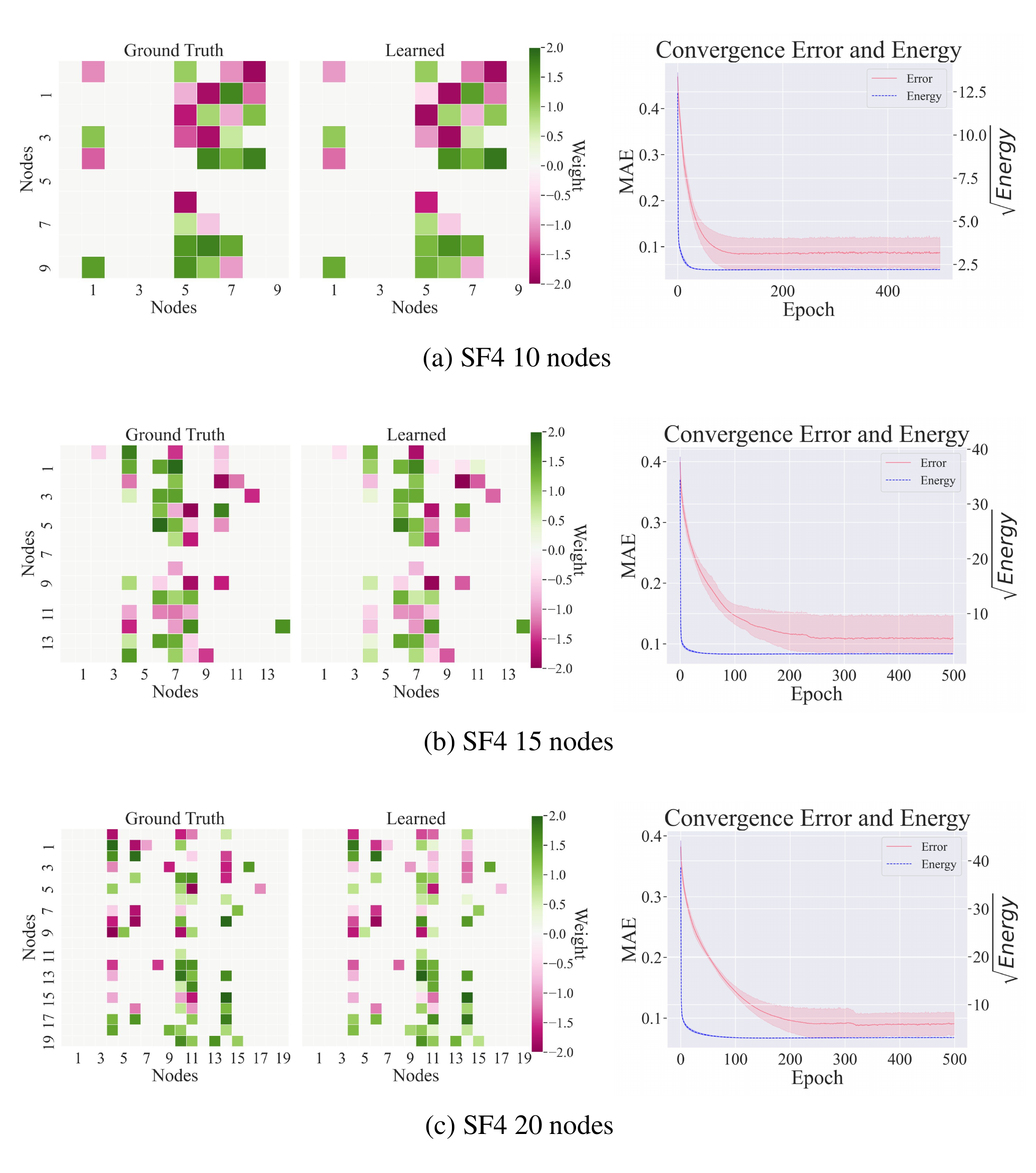}
\caption{Learned structures and convergence behavior (energy vs.\ MAE) for SF4 graphs of various complexity.}
\label{fig:SF4_cl}
\end{figure*}

First, we show results on learning the causal structure for the two most difficult graphs of our experiments, namely, ER2 and SF4 graphs, each with $N=20$. 
We see that PC graphs are able to learn good approximations of the ground truth weighted adjacency matrix of complex random graphs. This is depicted in the left columns of Figs.~\ref{fig:ER2_cl} and~\ref{fig:SF4_cl}, respectively. The right columns in Figs.~\ref{fig:ER2_cl} and~\ref{fig:SF4_cl} shows how the MAE decreases as the predictive coding objective converges. In all cases, we observe that while the energy converges early on, the causal discovery performance (MAE) keeps improving. Second, in Table~\ref{tab:table_CD}, we compare our method against established and recent causal discovery algorithms. In contrast to the baselines, our method consistently exhibits a good performance across various graph structures and does not deteriorate strongly for graphs of varying complexity by maintaining its causal structure learning abilities despite irregular node degree distributions and an increasing number of edges and nodes. This is reflected in the high accuracy metrics (F1) and low structural hamming distance (SHD) obtained with our method.  From our experiments, we observed that our method performs very well on scale-free graphs different to most of the benchmarks, which struggle on such random graphs. To conclude, the causal structure learning experiments conducted demonstrate that our method can learn arbitrary DAG structures of varying characteristics and levels of complexity. The complexity is determined by factors such as the number of nodes in the graph and the degree distribution of each node. Furthermore, our method demonstrates a robust performance even in the face of challenges such as increasingly uneven degree distributions and growing node cardinality. This distinguishes our approach from the baseline methods, thereby further highlighting the effectiveness of our predictive coding framework for causal structure learning.

\newpage

\newpage

\begin{table*}[htbp]
\Large
\centering
\resizebox{\textwidth}{!}{%
\begin{tabular}{cccccccccc}
\toprule
model & $N$ & graph & FDR $\downarrow$ &  TPR $\uparrow$ &  FPR $\downarrow$ & SHD $\downarrow$ & NNZ - & F1 $\uparrow$ \\
    \midrule
    Ours & &  & 0.08 $\pm$ 0.04 & 0.92 $\pm$ 0.04 & 0.02 $\pm$ 0.01 & \textbf{0.80 $\pm$ 0.45} & 10.00 $\pm$ 0.00 & \textbf{0.92 $\pm$ 0.04} \\
    GES &  &  & 0.22 $\pm$ 0.08 & 0.88 $\pm$ 0.11 & 0.07 $\pm$ 0.03 & 3.00 $\pm$ 1.41 & 11.20 $\pm$ 0.45 & 0.68 $\pm$ 0.09 \\
    PC & 10 & ER1 &  0.13 $\pm$ 0.04 & 0.82 $\pm$ 0.04 & 0.03 $\pm$ 0.01 & 2.20 $\pm$ 0.45 & 9.40 $\pm$ 0.55 & 0.77 $\pm$ 0.03 \\    
    ICALiNGAM &  &  & 0.25 $\pm$ 0.15 & 0.86 $\pm$ 0.11 & 0.09 $\pm$ 0.05 & 3.20 $\pm$ 2.05 & 11.60 $\pm$ 1.14 & 0.80 $\pm$ 0.13 \\
    NOTEARS &  &  & 0.08 $\pm$ 0.08 & 0.90 $\pm$ 0.07 & 0.02 $\pm$ 0.02 & 1.20 $\pm$ 0.84 & 9.80 $\pm$ 0.84 & 0.91 $\pm$ 0.07 \\
    \midrule
    Ours &  &  & 0.02 $\pm$ 0.04 & 0.94 $\pm$ 0.04 & 0.02 $\pm$ 0.04 & \textbf{1.40 $\pm$ 0.89} & 19.20 $\pm$ 1.30 & \textbf{0.96 $\pm$ 0.03} \\
    GES &  &  & 0.86 $\pm$ 0.03 & 0.26 $\pm$ 0.05 & 1.30 $\pm$ 0.02 & 33.60 $\pm$ 0.55 & 37.60 $\pm$ 1.14 & 0.17 $\pm$ 0.03 \\
    PC & 10 & ER2 & 0.47 $\pm$ 0.09 & 0.45 $\pm$ 0.08 & 0.32 $\pm$ 0.06 & 13.20 $\pm$ 1.48 & 17.00 $\pm$ 0.71 & 0.47 $\pm$ 0.08 \\
    ICALiNGAM &  &  & 0.31 $\pm$ 0.14 & 0.74 $\pm$ 0.12 & 0.27 $\pm$ 0.13 & 9.20 $\pm$ 4.76 & 21.60 $\pm$ 1.14 & 0.71 $\pm$ 0.13 \\
    NOTEARS &  &  & 0.14 $\pm$ 0.00 & 0.90 $\pm$ 0.00 & 0.12 $\pm$ 0.00 & 4.00 $\pm$ 0.00 & 21.00 $\pm$ 0.00 & 0.88 $\pm$ 0.00 \\
    \midrule % 
    Ours &  &  & 0.03 $\pm$ 0.04 & 0.99 $\pm$ 0.03 & 0.00 $\pm$ 0.01 & \textbf{0.60 $\pm$ 0.89} & 15.20 $\pm$ 0.45 & \textbf{0.98 $\pm$ 0.03} \\
    GES & &  & 0.22 $\pm$ 0.14 & 0.83 $\pm$ 0.10 & 0.04 $\pm$ 0.03 & 5.00 $\pm$ 3.00 & 16.00 $\pm$ 1.00 & 0.80 $\pm$ 0.12 \\
    PC & 15 & ER1 & 0.16 $\pm$ 0.05 & 0.85 $\pm$ 0.03 & 0.03 $\pm$ 0.01 & 3.40 $\pm$ 0.89 & 15.20 $\pm$ 0.45 & 0.78 $\pm$ 0.03 \\
    ICALiNGAM &  &  & 0.30 $\pm$ 0.06 & 0.77 $\pm$ 0.09 & 0.06 $\pm$ 0.01 & 5.40 $\pm$ 1.14 & 16.60 $\pm$ 1.52 & 0.73 $\pm$ 0.07 \\
    NOTEARS &  &  & 0.06 $\pm$ 0.08 & 0.89 $\pm$ 0.10 & 0.01 $\pm$ 0.01 & 2.00 $\pm$ 2.00 & 14.20 $\pm$ 0.84 & 0.92 $\pm$ 0.09 \\
    \midrule
    Ours &  &  & 0.20 $\pm$ 0.06 & 0.89 $\pm$ 0.04 & 0.09 $\pm$ 0.03 & \textbf{8.80 $\pm$ 3.11} & 33.60 $\pm$ 1.52 & \textbf{0.84 $\pm$ 0.05} \\
    GES &  &  & 0.56 $\pm$ 0.08 & 0.77 $\pm$ 0.07 & 0.40 $\pm$ 0.10 & 31.60 $\pm$ 7.70 & 52.80 $\pm$ 5.72 & 0.54 $\pm$ 0.08 \\
    PC & 15 & ER2  & 0.62 $\pm$ 0.07 & 0.37 $\pm$ 0.06 & 0.24 $\pm$ 0.03 & 33.40 $\pm$ 3.05 & 28.80 $\pm$ 1.48 & 0.37 $\pm$ 0.07 \\
    ICALiNGAM &  &  & 0.38 $\pm$ 0.08 & 0.75 $\pm$ 0.08 & 0.18 $\pm$ 0.04 & 17.60 $\pm$ 3.65 & 36.20 $\pm$ 2.49 & 0.68 $\pm$ 0.07 \\  
    NOTEARS &  &  & 0.17 $\pm$ 0.03 & 0.79 $\pm$ 0.02 & 0.07 $\pm$ 0.01 & 9.80 $\pm$ 0.84 & 28.60 $\pm$ 0.89 & 0.81 $\pm$ 0.02 \\
    \midrule
    Ours & &  & 0.25 $\pm$ 0.04 & 0.99 $\pm$ 0.02 & 0.04 $\pm$ 0.01 & \textbf{6.80 $\pm$ 1.48} & 26.60 $\pm$ 1.14 & \textbf{0.80 $\pm$ 0.04} \\
    GES &  &  & 0.43 $\pm$ 0.12 & 0.74 $\pm$ 0.13 & 0.07 $\pm$ 0.02 & 14.90 $\pm$ 6.47 & 26.30 $\pm$ 1.95 & 0.65 $\pm$ 0.13 \\
    PC & 20 & ER1 & 0.43 $\pm$ 0.08 & 0.61 $\pm$ 0.07 & 0.06 $\pm$ 0.01 & 13.80 $\pm$ 1.92 & 21.60 $\pm$ 1.14 & 0.54 $\pm$ 0.07 \\
    ICALiNGAM &  &  & 0.47 $\pm$ 0.05 & 0.72 $\pm$ 0.06 & 0.08 $\pm$ 0.01 & 14.60 $\pm$ 2.19 & 27.20 $\pm$ 1.92 & 0.61 $\pm$ 0.05 \\
    NOTEARS &  &  & 0.23 $\pm$ 0.08 & 0.79 $\pm$ 0.07 & 0.03 $\pm$ 0.01 & 8.80 $\pm$ 2.86 & 20.60 $\pm$ 0.55 & 0.78 $\pm$ 0.07 \\
    \midrule
    Ours &  &  & 0.12 $\pm$ 0.08 & 0.94 $\pm$ 0.02 & 0.04 $\pm$ 0.02 & \textbf{6.20 $\pm$ 3.83} & 43.00 $\pm$ 2.83 & \textbf{0.91 $\pm$ 0.05} \\
    GES &  &  & 0.70 $\pm$ 0.07 & 0.64 $\pm$ 0.11 & 0.42 $\pm$ 0.10 & 65.80 $\pm$ 13.57 & 88.80 $\pm$ 11.90 & 0.40 $\pm$ 0.09 \\
    PC & 20 & ER2 &  0.65 $\pm$ 0.04 & 0.37 $\pm$ 0.06 & 0.18 $\pm$ 0.01 & 44.00 $\pm$ 1.41 & 41.80 $\pm$ 2.39 & 0.36 $\pm$ 0.05 \\
    ICALINGAM &  &  & 0.34 $\pm$ 0.10 & 0.80 $\pm$ 0.06 & 0.11 $\pm$ 0.04 & 19.60 $\pm$ 6.47 & 49.20 $\pm$ 4.21 & 0.72 $\pm$ 0.08 \\
    NOTEARS &  &  & 0.15 $\pm$ 0.05 & 0.91 $\pm$ 0.02 & 0.04 $\pm$ 0.02 & 9.40 $\pm$ 1.95 & 42.80 $\pm$ 2.59 & 0.88 $\pm$ 0.03 \\
    \bottomrule
\end{tabular}%
}     
\caption{Comparison against established structure learning algorithms on various accuracy metrics for ER/SF graphs of increasing complexity. Mean and standard deviation calculated over 10 seeds.}
\label{tab:table_CD}
\end{table*}

\begin{table*}[htbp]
\Large
\ContinuedFloat % Indicates this is a continuation of the previous table
\centering
% Your second part of the table goes here
\resizebox{\textwidth}{!}{%
\begin{tabular}{cccccccccc}
\toprule
model & $N$ & graph & FDR $\downarrow$ &  TPR $\uparrow$ &  FPR $\downarrow$ & SHD $\downarrow$ & NNZ - & F1 $\uparrow$ \\
    \midrule
    Ours & & & 0.03 $\pm$ 0.09 & 0.99 $\pm$ 0.04 & 0.02 $\pm$ 0.07 & \textbf{0.70 $\pm$ 2.21} & 17.40 $\pm$ 1.26 & \textbf{0.98 $\pm$ 0.07} \\
    GES &  &  & 0.39 $\pm$ 0.02 & 0.89 $\pm$ 0.03 & 0.35 $\pm$ 0.05 & 9.80 $\pm$ 1.30 & 25.00 $\pm$ 1.73 & 0.69 $\pm$ 0.02 \\
    PC & 10 & SF2 & 0.24 $\pm$ 0.03 & 0.69 $\pm$ 0.03 & 0.14 $\pm$ 0.02 & 7.40 $\pm$ 0.55 & 15.60 $\pm$ 0.55 & 0.72 $\pm$ 0.02 \\
    ICALiNGAM &  &  & 0.40 $\pm$ 0.09 & 0.79 $\pm$ 0.07 & 0.33 $\pm$ 0.09 & 10.80 $\pm$ 2.77 & 22.60 $\pm$ 1.67 & 0.68 $\pm$ 0.08 \\    
    NOTEARS &  &  & 0.00 $\pm$ 0.00 & 0.82 $\pm$ 0.00 & 0.00 $\pm$ 0.00 & 3.00 $\pm$ 0.00 & 14.00 $\pm$ 0.00 & 0.90 $\pm$ 0.00 \\
    \midrule
    Ours &  &  & 0.08 $\pm$ 0.08 & 0.94 $\pm$ 0.06 & 0.17 $\pm$ 0.19 & \textbf{3.50 $\pm$ 3.78} & 30.80 $\pm$ 1.87 & \textbf{0.93 $\pm$ 0.07} \\   
    GES &  &  & 0.60 $\pm$ 0.05 & 0.56 $\pm$ 0.07 & 1.69 $\pm$ 0.12 & 27.00 $\pm$ 2.35 & 42.20 $\pm$ 0.45 & 0.44 $\pm$ 0.05 \\
    PC & 10 & SF4 & 0.18 $\pm$ 0.08 & 0.59 $\pm$ 0.06 & 0.25 $\pm$ 0.12 & 14.60 $\pm$ 2.41 & 21.40 $\pm$ 1.52 & 0.68 $\pm$ 0.06 \\
    ICALiNGAM &  &  & 0.21 $\pm$ 0.06 & 0.83 $\pm$ 0.05 & 0.45 $\pm$ 0.15 & 9.40 $\pm$ 3.21 & 31.80 $\pm$ 1.64 & 0.81 $\pm$ 0.06 \\
    NOTEARS &  &  & 0.03 $\pm$ 0.05 & 0.83 $\pm$ 0.09 & 0.05 $\pm$ 0.07 & 5.40 $\pm$ 3.29 & 25.80 $\pm$ 1.64 & 0.89 $\pm$ 0.07 \\
    \midrule
    Ours &  &  & 0.02 $\pm$ 0.02 & 0.98 $\pm$ 0.02 & 0.01 $\pm$ 0.01 & \textbf{0.60 $\pm$ 0.55} & 27.00 $\pm$ 0.00 & \textbf{0.98 $\pm$ 0.02} \\  
    GES &  &  & 0.27 $\pm$ 0.02 & 0.99 $\pm$ 0.02 & 0.13 $\pm$ 0.01 & 10.00 $\pm$ 1.00 & 36.80 $\pm$ 0.84 & 0.79 $\pm$ 0.01 \\
    PC & 15 & SF2 & 0.15 $\pm$ 0.08 & 0.76 $\pm$ 0.03 & 0.05 $\pm$ 0.03 & 9.40 $\pm$ 2.70 & 24.00 $\pm$ 1.41 & 0.77 $\pm$ 0.04 \\
    ICALiNGAM &  &  & 0.36 $\pm$ 0.11 & 0.87 $\pm$ 0.07 & 0.17 $\pm$ 0.07 & 15.20 $\pm$ 7.26 & 37.20 $\pm$ 3.83 & 0.74 $\pm$ 0.10 \\
    NOTEARS &  &  & 0.02 $\pm$ 0.02 & 0.97 $\pm$ 0.02 & 0.01 $\pm$ 0.01 & 1.40 $\pm$ 0.89 & 26.80 $\pm$ 0.45 & 0.97 $\pm$ 0.02 \\
    \midrule
    Ours &  &  & 0.10 $\pm$ 0.06 & 0.94 $\pm$ 0.05 & 0.09 $\pm$ 0.06 & \textbf{7.20 $\pm$ 4.55} & 52.20 $\pm$ 1.79 & \textbf{0.92 $\pm$ 0.05} \\
    GES &  &  & 0.47 $\pm$ 0.09 & 0.82 $\pm$ 0.10 & 0.67 $\pm$ 0.16 & 39.40 $\pm$ 11.04 & 77.60 $\pm$ 4.39 & 0.62 $\pm$ 0.09 \\
    PC & 15 & SF4 & 0.52 $\pm$ 0.06 & 0.29 $\pm$ 0.04 & 0.28 $\pm$ 0.03 & 45.40 $\pm$ 2.51 & 30.00 $\pm$ 0.71 & 0.36 $\pm$ 0.05 \\
    ICALiNGAM &  &  & 0.44 $\pm$ 0.09 & 0.69 $\pm$ 0.09 & 0.50 $\pm$ 0.11 & 36.40 $\pm$ 7.83 & 62.00 $\pm$ 3.39 & 0.62 $\pm$ 0.09 \\  
    NOTEARS &  &  & 0.12 $\pm$ 0.03 & 0.83 $\pm$ 0.08 & 0.11 $\pm$ 0.02 & 13.20 $\pm$ 4.92 & 47.40 $\pm$ 2.79 & 0.85 $\pm$ 0.06 \\
    \midrule
    Ours & &  & 0.10 $\pm$ 0.14 & 0.96 $\pm$ 0.05 & 0.03 $\pm$ 0.04 & \textbf{5.40 $\pm$ 7.40} & 40.20 $\pm$ 4.38 & \textbf{0.93 $\pm$ 0.10} \\
    GES &  &  & 0.30 $\pm$ 0.12 & 0.96 $\pm$ 0.02 & 0.10 $\pm$ 0.05 & 16.00 $\pm$ 7.28 & 51.20 $\pm$ 6.98 & 0.77 $\pm$ 0.08 \\
    PC & 20 & SF2 &  0.26 $\pm$ 0.07 & 0.70 $\pm$ 0.06 & 0.06 $\pm$ 0.02 & 17.80 $\pm$ 4.38 & 34.80 $\pm$ 1.48 & 0.68 $\pm$ 0.06 \\
    ICALiNGAM &  &  & 0.34 $\pm$ 0.17 & 0.87 $\pm$ 0.06 & 0.12 $\pm$ 0.08 & 19.80 $\pm$ 13.01 & 50.80 $\pm$ 9.88 & 0.75 $\pm$ 0.13 \\
    NOTEARS & & & 0.23 $\pm$ 0.07 & 0.88 $\pm$ 0.04 & 0.06 $\pm$ 0.02 & 12.80 $\pm$ 5.76 & 42.20 $\pm$ 2.28 & 0.82 $\pm$ 0.05 \\
    \midrule
    Ours &  &  & 0.06 $\pm$ 0.05 & 0.98 $\pm$ 0.03 & 0.04 $\pm$ 0.03 & \textbf{5.60 $\pm$ 5.50} & 73.00 $\pm$ 1.73 & \textbf{0.96 $\pm$ 0.04} \\
    GES &  &  & 0.28 $\pm$ 0.03 & 0.94 $\pm$ 0.04 & 0.21 $\pm$ 0.02 & 26.60 $\pm$ 3.13 & 91.40 $\pm$ 1.52 & 0.79 $\pm$ 0.03 \\
    PC & 20 & SF4 & 0.38 $\pm$ 0.04 & 0.34 $\pm$ 0.02 & 0.12 $\pm$ 0.01 & 54.60 $\pm$ 2.70 & 38.80 $\pm$ 1.30 & 0.44 $\pm$ 0.03 \\
    ICALiNGAM &  &  & 0.37 $\pm$ 0.13 & 0.73 $\pm$ 0.05 & 0.27 $\pm$ 0.13 & 45.20 $\pm$ 17.25 & 83.40 $\pm$ 12.46 & 0.67 $\pm$ 0.10 \\
    NOTEARS &  &  & 0.18 $\pm$ 0.01 & 0.81 $\pm$ 0.02 & 0.10 $\pm$ 0.01 & 22.80 $\pm$ 1.64 & 69.40 $\pm$ 0.55 & 0.82 $\pm$ 0.02 \\
\bottomrule
\end{tabular}%
} 

\caption[]{Comparison against established structure learning algorithms on various accuracy metrics for ER/SF graphs of increasing complexity. Mean and standard deviation calculated over 10 seeds. (cont.)} % The empty square brackets [] ensure the table is treated as a continuation
\label{tab:table_CD_cont} % Optional: different label if you need to refer specifically to this part of the table
\end{table*}

\vspace{-1ex}
\subsection{Classification}
\vspace{-1ex}
In the main body of this work, we showed that pruning unnecessary connections in a complete graph results in a hierarchical structure with improved performance. In this section, we provide further details to reproduce our results.

The classification experiments are performed on the MNIST, FashionMNIST, and 2-MNIST datasets. The latter is obtained by pairing the image $\hat{\x}$ in each sample $(\hat{\x}, \hat{\y})$ of the MNIST dataset with a new digit image $\hat{\x}'$ sampled uniformly from the dataset (while maintaining the train and test splitting intact). Thus, a data point of the 2-MNIST dataset consists of the tuple $(\hat{\x}, \hat{\x}', \hat{\y})$. 

We consider a predictive coding graph with 6 nodes, one of dimension $784$ (input dimension), one of dimension $10$ (output dimension), and 4 hidden nodes of dimension $d$. In the case of 2-MNIST, we have two nodes of dimension 784, and only three of dimension $d$. The results reported in this work were obtained with $d = 128$. During training, we used a batch size of $512$ and $T=32$. To start from a complete graph, as the one defined in Section \ref{sec:ci}, we define a fully connected layer $f_{i,j}$, with $gelu$ activation function, going from node $i$ to node $j$ for each ordered pair of nodes $(i,j)$. The output of each layer $f_{i,j}$ is then multiplied by a scaling factor $a_{i,j}$ that determines the strength of the connection from node $i$ to node $j$. Together, the factors $a_{i,j}$ determine the adjacency matrix $A$. To enforce sparse connectivity and prune unnecessary edges, we add to the matrix $A$ the $L1$ regularizer $l(A)$. Consequently, the loss function to optimize becomes $\mathcal{L} = F + \omega \cdot l(A)$, where $\omega$ is a weighting factor.

\begin{figure}
  \begin{center}
  \vspace{-0.3cm}
\includegraphics[width=0.7\columnwidth]{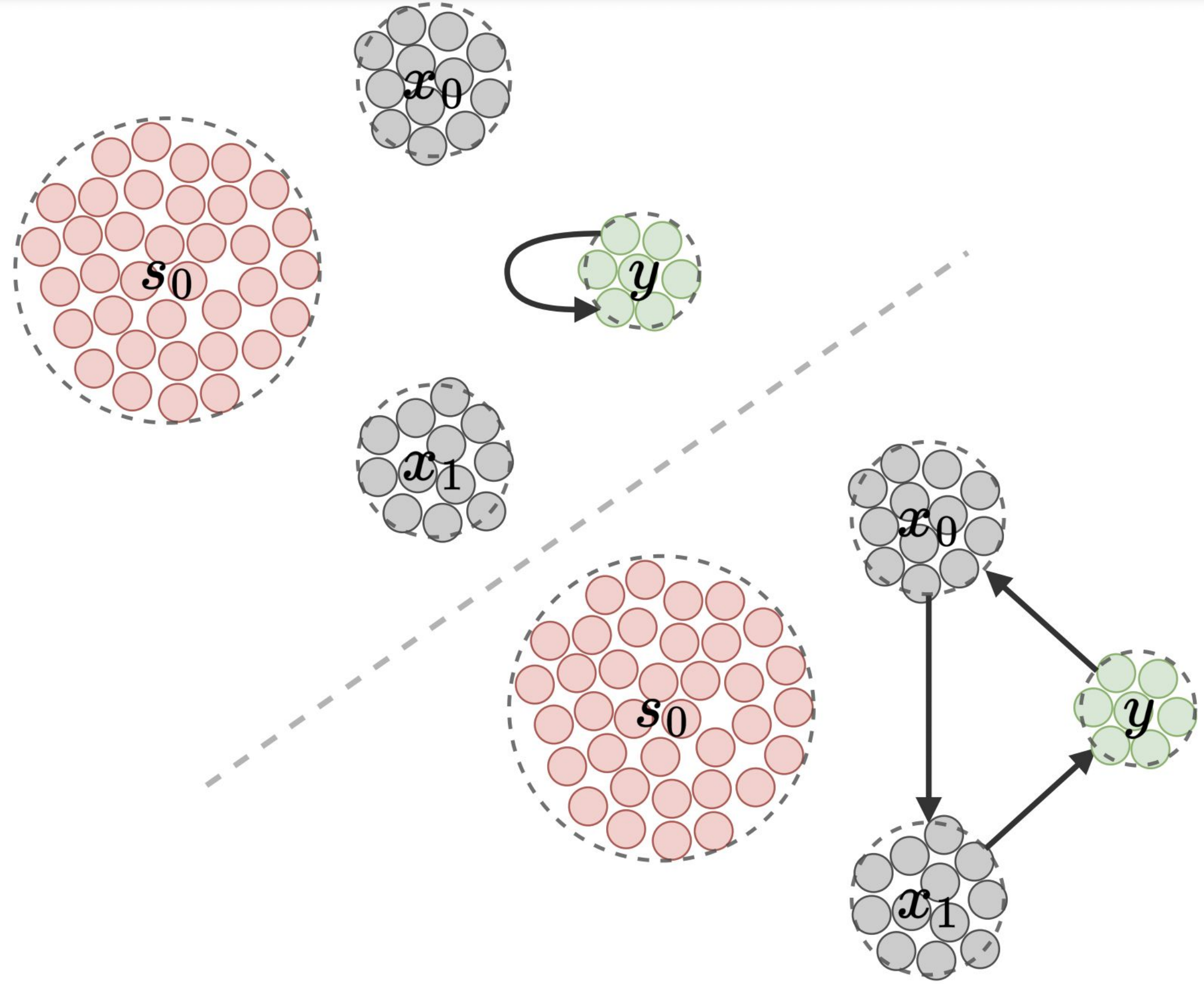}
  \end{center}
  \caption{Examples of degenerate networks, where the label predicts itself either via self loops, or via cycles.}
  \label{fig:degenerate}
\vspace{-0.4cm}
\end{figure}

\textbf{Degenerate Example.}
We start our discussion by showing a degenerate example, which arises when we do not use either negative examples, or a prior that forces an acyclic structure, but only the prior $l(A)$ which enforces sparsity. In this case, the modes is unable to learn the causal dependency between input and output, and converges towards a degenerate structure, where each output node predicts itself via a cyclic structure, which can be a self loop, or a closed loop with length larger than one. As each node either predicts itself or is unused, the total variational free energy of the network is going to be close to $0$, despite the network being randomly guessing the output. An example of such structures is provided in  Fig.~\ref{fig:sl_exp}. This shows the importance of additional methods, that force the network to be aware of the causal dependency between the input and the output. We now test the two proposed methods: the acyclic prior, and the use of negative examples.

To overcome this, we propose two different methods:
\begin{enumerate}
    \item \textit{force an acyclic structure}, by adding to the loss function the regulariser $h(A) = tr(exp(A \times A))$ introduced in \citep{zheng2018dags}. We weight $h(A)$ by a scalar $\eta$.
    \item \textit{force a connection between input nodes and output nodes}, by introducing negative samples in the training dataset: with probability $p_{ns}$ we sample randomly a new label $\hat{y}_{ns}$. We modify the energy function:
    \begin{equation}
        \hat{F} = \textstyle \sum_{i \neq i_\y} \| \x_i - \u_i \|^2 + (\|\x_{i_y} - \mu_{i_y}\|^2 - k)^2,
    \end{equation}
    where $i_y$ is the index of the node fixed to $\hat{\y}$ and $k$ the new energy target. We set $k = 0$ for positive samples and $k > 0$ for negative samples. With negative samples, the output node cannot simply learn to predict itself, as the energy would be non-zero for negative samples, for which the energy target is $k > 0$.
\end{enumerate}
\textbf{Discussion.}
Both methods produce hierarchical structures that achieve a better performance than the original complete graph. Method (1) has the disadvantage of introducing an inductive bias in the architecture by completely removing loops and requiring a complex balance between the $\omega$ and $\eta$ parameters, as they affect each other. On the other hand, method (2), despite overcoming these issues, seems more brittle with respect of the choice of hyperparameters and produces a smaller variety of networks.
The value of $\omega$ determines the overall network structure. For method (1), we observed a wide range of possible output structures (e.g., using zero or multiple hidden-nodes, in parallel or in sequence, with and with-out skip connections) depending on the chosen $\omega$. The best accuracy, however, was always achieved with a structure equivalent to a hierarchical neural network with two fully connected layers as shown in Figure~\ref{fig:sl_exp}. For method (2), instead, the only non-degenerate possibilities were either a complete graph (for low $\omega$ values), the optimal 2-layer network, or a network with no edges (for high $\omega$ values). A possible future research direction could be aiming at combining the two methods to overcome their respective limitations. Table~\ref{tab:classification_params} reports the best hyperparameters for each method and dataset.
\begin{table*}[t]
    \centering
    \begin{tabular}{c|c|cccccc}
        \toprule
        Dataset & Method & $\gamma$ & $\alpha$ & $\beta$ & $\omega$ & $\eta$ & $p_{ns}$ \\
        \midrule
        MNIST & DAG & $0.5$ & $4e-05$ & $2e-05$ & $8e-$4 & $40.0$ & - \\
        Fashion-MNIST & DAG & $0.3$ & $3e-05$ & $5e-05$ & $1e-3$ & $20.0$ & - \\
        2-MNIST & DAG & $0.5$ & $4e-05$ & $2e-05$ & $8e-4$ & $40.0$ & - \\
        \midrule
        MNIST & NS & $0.8$ & $1e-04$ & $8e-05$ & $0.05$ & - &$ 0.1$ \\
        Fashion-MNIST & NS &$ 0.5$ & $1e-04$ & $8e-05$ & $0.05$ & - & $0.2$\\
        2-MNIST & NS & $0.8$ & $1e-04$ & $8e-05$ & $0.05$ & - & $0.1$\\
        \bottomrule
    \end{tabular}

    \medskip 
    \caption{Hyperparameters used to obtain the results reported in Fig.~\ref{fig:sl_exp}. \textit{DAG} (directed acyclic graphs) refers to method (1), while \textit{NS} (negative samples) to method (2). The accuracy obtained on 2-MNIST is similar to the one obtained for MNIST. $k$ was set to $1.0$ for negative samples. A weight decay of $0.001$ was applied to the node values during the \textit{NS} experiments.}
    \label{tab:classification_params}
\end{table*}

\section{End-to-end Causal Learning}

The goal of the presented experiments so far was to show that a causal predictive coding network can solve both tasks of causality:

\begin{itemize}
    \item Given observational data, perform unsupervised causal structure learning of the (weighted) adjacency matrix that represents the data generating SCM.
    \item Given observational data and causal structure, perform inference of associational, interventional, and counterfactual distributions to answer causal queries. 
\end{itemize}

Therefore, this section is motivated by studying the capability of our proposed method to combine both tasks into a single framework. We use the same PC graph to conduct structure learning for the common graphs used in the causal inference experiments (\emph{chain, collider, confounder, fork, mediator, butterfly bias, M-bias}) using only observational data generated by the corresponding SCM. 
Given that our method is able to (i) discover complex causal structures for random DAGs in Section~\ref{sec:sl} and (ii) correctly answer causal queries for common DAG structures in Section~\ref{sec:ci}, we hypothesize that our method should be able to discover graph structures used in Section~\ref{sec:ci} without prior knowledge.
The motivation behind this approach is that in the real-world, true \emph{causal structures are rarely known} and often only observational data from an \emph{unknown SCM} is available.
In the following, we perform causal structure learning for the graphs used in the causal inference experiments of Section~\ref{sec:ci}. The procedure is as follows: 
First, given observational data, we learn the causal structure of the underlying SCM that generated the observed data. More specifically, we start with a fully connected PC graph and prune unnecessary node connections using sparsity and acyclicity constraints with subsequent thresholding as described in Appendix~\ref{sec:app_cg}.
Second, given the observational data and the discovered causal structure, we learn the SCM parameters including an approximation of the parameters of each node's exogenous distribution $\mathbf{u}_i$. To be more specific, once the causal structure is discovered, we modify the PC graph by including one exogenous node, $\mathbf{u}_i$, for each endogenous variable, $\x_i$, into our PC graph. Augmenting the PC graph with exogenous nodes enables us to learn the distribution of each exogenous node, which is crucial in the SCM framework, as exogenous variables are essential for conducting counterfactual inference.
This procedure provides us with a simple and closed form end-to-end causal inference engine.
Using a single PC model with no pipelines, enables us to (1) discover the adjacency matrix of the SCM and to (2) answer causal queries on any of the three levels of Pearl's ladder of causation (\citeyear{pearl2009causality}). We report results for the structure learning step in Table~\ref{tab:End_to_end_causality_table}.

\begin{table*}[htbp]
\centering
\resizebox{\textwidth}{!}{%
%\begin{tabular}{lrlllllllll}
\begin{tabular}{ccccccccccc}
\toprule
graph & $N$ & MAE $\downarrow$ & FDR $\downarrow$ & TPR $\uparrow$ & FPR $\downarrow$ & SHD $\downarrow$ & NNZ - & F1 $\uparrow$ \\
\midrule
butterfly & 5 & 0.02 $\pm$ 0.01 & 0.00 $\pm$ 0.00 & 1.00 $\pm$ 0.00 & 0.00 $\pm$ 0.00 & 0.00 $\pm$ 0.00 & 6.00 $\pm$ 0.00 & 1.00 $\pm$ 0.00 \\
M & 5 & 0.03 $\pm$ 0.01 & 0.04 $\pm$ 0.09 & 1.00 $\pm$ 0.00 & 0.03 $\pm$ 0.07 & 0.20 $\pm$ 0.45 & 4.20 $\pm$ 0.45 & 0.98 $\pm$ 0.05 \\
chain & 3 & 0.01 $\pm$ 0.00 & 0.00 $\pm$ 0.00 & 1.00 $\pm$ 0.00 & 0.00 $\pm$ 0.00 & 0.00 $\pm$ 0.00 & 2.00 $\pm$ 0.00 & 1.00 $\pm$ 0.00 \\
confounder & 3 & 0.22 $\pm$ 0.12 & 0.27 $\pm$ 0.15 & 0.73 $\pm$ 0.15 & 0.80 $\pm$ 0.45 & 0.80 $\pm$ 0.45 & 3.00 $\pm$ 0.00 & 0.73 $\pm$ 0.15 \\
collider & 3 & 0.01 $\pm$ 0.00 & 0.00 $\pm$ 0.00 & 1.00 $\pm$ 0.00 & 0.00 $\pm$ 0.00 & 0.00 $\pm$ 0.00 & 2.00 $\pm$ 0.00 & 1.00 $\pm$ 0.00 \\
fork & 3 & 0.01 $\pm$ 0.01 & 0.00 $\pm$ 0.00 & 1.00 $\pm$ 0.00 & 0.00 $\pm$ 0.00 & 0.00 $\pm$ 0.00 & 2.00 $\pm$ 0.00 & 1.00 $\pm$ 0.00 \\
mediator & 3 & 0.46 $\pm$ 0.33 & 0.27 $\pm$ 0.15 & 0.73 $\pm$ 0.15 & 0.80 $\pm$ 0.45 & 0.80 $\pm$ 0.45 & 3.00 $\pm$ 0.00 & 0.73 $\pm$ 0.15 \\
\bottomrule
\end{tabular}%
}

\medskip 
\caption{End-to-end causality engine: Causal predictive coding for discovery of DAGs based on observational data only. Numbers are reported over 5 different runs.}
\label{tab:End_to_end_causality_table}
\end{table*}

Despite the graphs being very different, the experimental results show that our method performs well in causal discovery for most common graphs despite  using the same hyperparameters and no hyperparameter search, even though the graphs are very different. We do not show the causal inference results again,  because the results for learning associational, interventional, and counterfactual distributions and the distribution of the exogenous noise variables in SCM remained the same. The discovered causal structures are consistent with the adjacency matrices used as prior knowledge in Section~\ref{sec:ci}. Our method is able to solve both causality tasks without prior knowledge of any graph structures. We showed how our causal predictive coding framework can be used in an end-to-end unsupervised causal inference pipeline similar to \citep{geffner2022deep} but without the need of complex neural networks. Thus, our proposed causal predictive coding maintains transparency and interpretability despite good performance.

\end{document}